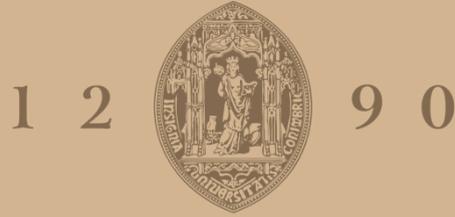

UNIVERSIDADE Ð
COIMBRA

Diana Maria Conceição Mortágua

# Improving Annotator Selection in Active Learning Using a Mood and Fatigue-Aware Recommender System

Dissertation in the context of the Masters in Data Science and Engineering, advised by Professor Doctor Luís Macedo and Professor Doctor Amílcar Cardoso and presented to the Department of Informatics Engineering of the Faculty of Sciences and Technology of the University of Coimbra.

July 2025

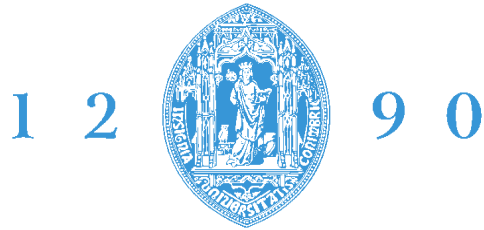

DEPARTAMENTO DE
ENGENHARIA INFORMÁTICA
FACULDADE DE
CIÊNCIAS E TECNOLOGIA
UNIVERSIDADE Đ
COIMBRA

Diana Maria Conceição Mortágua

# Melhorando a seleção de anotadores em Active Learning usando um Sistema de Recomendação Sensível à Disposição e à Fadiga

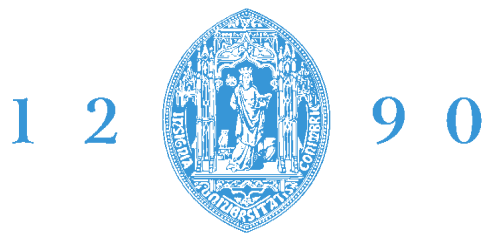

DEPARTAMENTO DE
ENGENHARIA INFORMÁTICA
FACULDADE DE
CIÊNCIAS E TECNOLOGIA
UNIVERSIDADE Ð
COIMBRA

Diana Maria Conceição Mortágua

# Improving Annotator Selection in Active Learning Using a Mood and Fatigue-Aware Recommender System

Dissertation in the context of the Masters in Data Science and Engineering, advised by Professor Doctor Luís Macedo and Professor Doctor Amílcar Cardoso and presented to the Department of Informatics Engineering of the Faculty of Sciences and Technology of the University of Coimbra.

July 2025

# Acknowledgements

Foremost, I want to thank my family for everything they did so that I could study and conclude this project. All the love, encouragement and strength they gave me allowed me to fight for my dreams. Thank you, Mom and Dad, for believing in me and sacrificing in order to let me reach my goals. You are my biggest pillars, who allow me to fall and get back up. Thank you, aunt Adelaide, for all the sweet treats, success wishes and the constant presence.

I also want to thank my advisors Professor Luís Macedo and Professor Amílcar Cardoso, who supported and guided me through the course of this project. Your advice and words of encouragement are deeply appreciated, and I am deeply grateful for the availability and guidance provided.

Finally, I want to thank my friends and colleagues, who have shared laughs and wisdom with me. I will always cherish the moments we spent together. I would like to specially thank Leonor, Catarina, Rama, and Nuno for all the amazing times we have spent together. I feel incredibly thankful to have people like you in my life. You inspire me, and allow me to recharge and relax by creating memories together.

This work was supported by the Portuguese Recovery and Resilience Plan through project C645008882-00000055, Center for Responsible AI; and by national funds through FCT – Foundation for Science and Technology, I.P. (grant number UI/BD/153496/2022), within the scope of the project CISUC (UID/CEC/00326/2020).

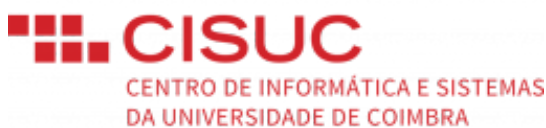 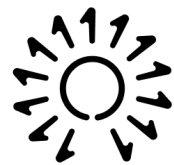

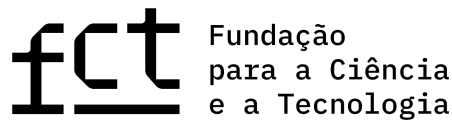

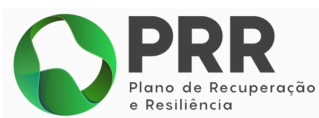 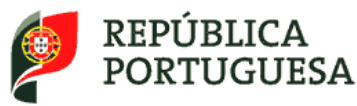 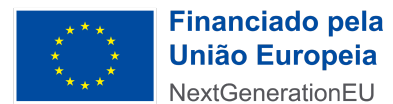

STATEMENT ON THE USE OF ARTIFICIAL INTELLIGENCE TOOLS

This document was slightly refined with the assistance of Large Language Models, such as ChatGPT, or similar tools (Grammarly), which helped check grammar, correct typos, and enhance clarity. Any Artificial Intelligence-generated text or content is clearly indicated within the document and is used only to a limited extent. The overall content and ideas remain solely the responsibility of the author.

# Abstract

This study centers on overcoming the challenge of selecting the best annotators for each query in Active Learning (AL), with the objective of minimizing misclassifications. AL recognizes the challenges related to cost and time when acquiring labeled data, and decreases the number of labeled data needed. Nevertheless, there is still the necessity to reduce annotation errors, aiming to be as efficient as possible, to achieve the expected accuracy faster [Munro, 2021]. While multiple strategies for query-annotator pairs selection consider the imperfect nature of human annotators, and try to overcome this problem (such as proactive AL [Bahle et al., 2016; Calma et al., 2016] and collaborative interactive learning [Calma et al., 2016, 2018a; Donmez et al., 2008; Donmez and Carbonell, 2010]), most do not consider internal factors that affect productivity. Some of these factors are mood, attention, motivation, and fatigue levels, which have been shown to affect productivity [Caldwell et al., 2019; Calma et al., 2016; Fredrickson and Branigan, 2005; Tenney, Poole, and Diener, 2016].

This work addresses this gap in the existing literature, by not only considering how the internal factors influence annotators (mood and fatigue levels) but also presenting a new query-annotator pair strategy, using a Knowledge-Based Recommendation System (RS). The RS ranks the available annotators, allowing to choose one or more to label the queried instance using their past accuracy values, and their mood and fatigue levels, as well as information about the instance queried. This work bases itself on existing literature on mood and fatigue influence on human performance, simulating annotators in a realistic manner, and predicting their performance with the RS.

The results show that considering past accuracy values, as well as mood and fatigue levels reduces the number of annotation errors made by the annotators, and the uncertainty of the model through its training, when compared to not using internal factors. Accuracy and F1-score values were also better in the proposed approach, despite not being as substantial as the aforementioned. The methodologies and findings presented in this study begin to explore the open challenge of human cognitive factors affecting AL.

# Keywords

# Sumário

Este estudo centra-se em superar o desafio de selecionar os melhores anotadores para cada consulta em Active Learning (AL), com o objetivo de minimizar as classificações incorretas. AL reconhece os desafios relacionados com o custo e o tempo ao adquirir dados rotulados, e diminui a quantidade de dados rotulados necessários. No entanto, existe ainda a necessidade de reduzir os erros de anotação, com o objetivo de ser o mais eficiente possível, para atingir a precisão esperada mais rapidamente [Munro, 2021]. Embora várias estratégias para a seleção de pares de consultas-anotadores considerem a natureza imperfeita dos anotadores humanos e tentem ultrapassar este problema (como a aprendizagem assistida proativa [Bahle et al., 2016; Calma et al., 2016] e a aprendizagem interativa colaborativa [ Calma et al., 2016, 2018a; Donmez et al., 2008; Donmez e Carbonell, 2010]), a maioria não considera fatores internos que afetam a produtividade. Alguns destes fatores são o humor, a atenção, a motivação e os níveis de cansaço, que demonstraram afetar a produtividade [Caldwell et al., 2019; Calma et al., 2016; Fredrickson e Branigan, 2005; [Tenney, Poole e Diener, 2016].

Este trabalho visa abordar esta lacuna na literatura existente, não só considerando como os fatores internos influenciam os anotadores, mas também apresentando uma nova estratégia de criação de pares query-anotador, utilizando um Knowledge-Based Recommendation System (RS). Este método ordena os anotadores disponíveis, o que permite selecionar um ou mais para rotular a instância, usando valores da sua precisão histórica, do seu humor e nível de fadiga, bem como informação da amostra a anotar. Este trabalho baseia-se na literatura existente sobre a influência do humor e do cansaço no trabalho de humanos, simulando anotadores de forma realista, e prevendo a sua precisão através do RS.

Os resultados mostram que considerar valores antigos de precisão, bem como o humor e nível de fadiga reduz o número de erros de anotação feitos pelos anotadores, e a incerteza do modelo durante o seu treino, quando comparado com não usar estes fatores internos. Os valores da precisão e do F1-score também melhoraram com a proposta apresentada, apesar de não tão substancialmente como aqueles acima mencionados. As metodologias e descobertas apresentadas neste estudo começam a explorar a lacuna de como fatores cognitivos humanos afetam AL.

The results show that considering past accuracy values, as well as mood and fatigue levels reduces the number of annotation errors made by the annotators, and the uncertainty of the model through its training when compared to not using internal factors. Accuracy and F1-score values were also better in the proposed approach, despite not being as substantial as the aforementioned. The methodologies and findings presented in this study begin to explore the open challenge of considering human cognitive factors in their performance on AL.

# Palavras-Chave

Active Learning, Sistemas de Recomendação, Fatores Cognitivos Humanos, Pares Query-Anotador

# Contents

# Acronyms

**AI** Artificial Intelligence

**AL** Active Learning

**BAmid** Best Alertness interval midpoint

**CBF** Content-Based Filtering

**CF** Collaborative Filtering

**HitL** Human in the Loop

**ML** Machine Learning

**MSFsc** Mid-Sleep on Free days corrected

**NA** Negative Affect

**PA** Positive Affect

**QBC** Query-by-Committee

**RS** Recommendation System

**SL** Supervised Learning

**UL** Unsupervised Learning

# List of Figures

# Chapter 1
# Introduction

This chapter starts by introducing the main motivation behind our work, as well as the problem statement and the research questions that propelled this study. It ends with the outline explaining the document's structure.

## 1.1 Motivation

Machine Learning (ML) models rely mostly on labeled datasets, where each data instance has its corresponding label correctly assigned. However, acquiring labeled data is difficult, as it is costly and time-consuming, and humans commonly annotate it [Munro, 2021, Settles, 2009]. Active Learning (AL) is an approach developed to address this issue. It aims to reduce annotations and misclassification costs while maintaining or even increasing performance [Munro, 2021; Settles, 2009]. Instead of annotating the complete dataset, AL selects some instances for which annotations will be more informative to the model, increasing its accuracy faster.

Traditional AL approaches base themselves on assumptions that limit their practical use due to their unrealistic nature. These are the presence of a singular omnipresent oracle whose labels are always correct, the constant cost of annotation across queries, and that each query asks for the label of each instance [Herde et al., 2021a; Settles, 2009]. However, it is common to use multiple imperfect annotators, whose performance is influenced by numerous internal and external factors. Furthermore, depending on the task and context, annotations have different complexity levels, and their cost fluctuates [Herde et al., 2021a; Munro, 2021]

To address these limitations, concepts have been developed, such as proactive AL [Bahle et al., 2016; Calma et al., 2016] and collaborative interactive learning [Calma et al., 2016, 2018a; Donmez et al., 2008; Donmez and Carbonell, 2010]. These approaches can be seen as real-world AL strategies that transform traditional AL into practices that can be methodically applied.

## 1.2 Problem Statement

Although proactive AL and collaborative interactive learning tackle most issues of traditional AL [Munro, 2021], research has not yet reached a consensus on the best way to address imperfect annotators, and how to be more effective in how to assign them to queried instances. Multiple internal factors affect annotators, making them more error-prone [Munro, 2021; Fredrickson and Branigan, 2005], and each annotator has different levels of knowledge in each field and a different cost of employment. Still, companies commonly just want to reduce all costs without compromising accuracy [Munro, 2021], making studies on real-world AL mostly focus on the annotators' cost and knowledge, not considering cognitive human factors [Herde et al., 2021a].

These internal human features (such as mood and fatigue level) in addition to annotators' reliability through past performance, could improve the model's performance [Fredrickson and Branigan, 2005]. This consideration acts as another layer of ensuring that the annotator is the most suited for that specific query at that moment.

This work researched those claims in an AL setting, using these internal factors (mood and fatigue levels) in a Recommendation System (RS) to select the best query-annotator pairs at a given moment. This way, one may investigate the effect of using internal factors on productivity of the workers and results of the AL task. The aim is to find a clear relationship that may be useful to use in strategies for selecting annotators in an AL setting. By controlling annotators' productivity on each work period, considering their fatigue and mood levels, this approach allows the AL model to achieve the desired performance level faster. This can save time and money for companies, as the selected annotators are expected to perform better, make fewer mistakes, and work more efficiently, saving money and time.

## 1.3 Research Questions

The main goal is to to investigate whether using RS to assign annotators to queries is an effective approach to getting query-annotator pairs, and to determine if using internal factors (mood and fatigue level) that may influence an annotator's performance improves the annotator's selection strategy in an AL setting.

Developing the main research questions, which can be expressed as follows, was essential to guide the research done in this study.

- **Research Question 1**: How effective is a Recommendation System (RS) in assigning annotators to queried instances in an AL setting, compared to traditional methods?

  To address this research question, the Knowledge-Based RS approach used to recommend annotators for each instance was compared to a traditional strategy that only considers annotators' past accuracy. Multiple metrics are used to compare different annotators' selection strategies, namely comparison of accuracy, F1-score, errors performed by the annotators, and learning curves.

- **Research Question 2**: Does introducing mood and fatigue levels of annotators as features in the Recommendation System improve the performance of the model compared to strategies that do not consider these internal factors?

  To address this question, a comparison is made by using the RS without accounting for mood and fatigue, and using the RS with those features. Comparison metrics such as accuracy, F1 score, errors performed by the annotators, and learning curves are used to assess which approach is the best.

- **Research Question 3**: How much should mood and fatigue levels be considered in the strategy used to select annotators to maximize the improvement of the AL model?

  To address this question, literature reviews are used when empirical data is available. If it is not, different values are tested to explore the results. Comparison metrics such as accuracy, F1 score, errors performed by the annotators, and learning curves are used to study different values.

- **Research Question 4**: Does introducing mood and fatigue levels of annotators as features in the RS improve the performance of the model compared to an optimized alternative?

To address a comparison is made by using the RS account for mood and fatigue, with using an optimized approach based on the simulated annotator's labeling process. Comparison metrics such as accuracy, F1 score, errors performed by the annotators, and learning curves are used to compare the approaches.

## 1.4 Thesis Outline

This document is structured into six chapters. The first chapter, which is the introductory chapter, outlines the topic and presents the main research questions. Chapter 2 covers fundamental knowledge essential for understanding the work conducted in this study. Chapter 3 summarizes the state-of-the-art of AL, and AL combined with RS, as well as the relationship between mood and fatigue on performance. Chapter 4 presents the methodology proposed in this study. Chapter 5 presents the results and their discussion, while Chapter 6 concludes the document by summarizing the key outcomes of the study and their implications.

# Chapter 2
# Background Knowledge

This chapter aims to deepen the reader's comprehension of the specific topics related to this work. It starts with explaining the field of Machine Learning, continuing to the presentation of Active Learning and Recommendation Systems. It finishes by giving background information on chronotypes and an overview of the presented concepts.

## 2.1 Machine Learning

Artificial Intelligence (AI) can be understood as a scientific field focused on understanding, analyzing, and replicating different kinds of intelligence and behaviors, as well as a branch of engineering that applies this understanding to create intelligent machines [Macedo, 2025]. Machine Learning (ML) is a subfield of AI that allows computers to learn from experience (often in the form of data) to improve the system's performance [Shinde & Shah, 2018]. The first step in the creation of ML was in 1948 when Tuning and Champernowne founded the world's first chess-playing computer program [Shinde & Shah, 2018]. Later, in 1951, Anthony Oettinger wrote the "response learning programme" (also known as "shopping programme") that simulated a kid's visit to a shop, which was the first AI program to incorporate learning [Shinde & Shah, 2018].

ML evolved since then, with the introduction of neuron studies, perceptrons, and multiple algorithms. Some of the major achievements were Support Vector Machines, in 1995, Adaboost, in 1997, Random Forest, in 2001, and the start of the ascent of Neural Networks in Deep Learning, in 2005. To date, these and many more ML algorithms, such as Logistic Regression, K-Nearest Neighbors, and Naive Bayes, are widely used across various applications [Shinde & Shah, 2018].

ML has many different applications that go from computer vision (like object recognition) and semantic analysis, to natural language processes and information retrieval [Shinde & Shah, 2018]. In addition to these, another key area of ML is prediction, which encompasses tasks such as recommendation and classification. Many ML algorithms can be used to classify different types of data (such as images, texts, and documents) for which it needs a lot of training data.

When a model is trained using labeled data, where each point has the corresponding correct label, we call it Supervised Learning (SL). The goal is for the model to generalize the training data, so that it correctly classifies new unlabeled instances. This type of learning is usually preferred to Unsupervised Learning (UL), where the algorithm just finds and learns patterns from the unlabeled data. It is estimated that supervised ML powers 90% of ML applications [Munro, 2021], but its downside is how hard it can be to obtain labeled data. While some labels, for instance, on a book rating website, can be acquired with little or no cost, it can be very time-consuming, difficult, or expensive to collect labeled data [Settles, 2009] (such as in speech recognition or document classification). This challenge motivates the creation of a process that allows models to perform better with less labeled data. One solution is Active Learning (AL), a subfield of ML (and consequently, of AI) that, when it uses human annotators, is a form of Human-in-the-Loop (HitL) [Munro, 2019].

HitL is a branch of AI that combines human and machine intelligence to create ML models. This collaboration aims to improve productivity and efficiency, using ML to assist with human tasks [Munro, 2021]. It promotes ethical decision-making and transparency in all steps by including the human in the process, such as through feedback and guidance. AL can be considered the linchpin of HitL approaches [Munro, 2021] as it identifies the most relevant data to present for feedback to the model in the form of annotations with labels.

## 2.2 Active Learning

An ML model may need thousands of labeled data instances to be trained. They can be difficult to obtain due to the time required to label them or the annotation cost itself (commonly done by humans). AL addresses this problem by aiming to reduce the amount of annotations needed for training and misclassification costs, while maintaining or improving the model's performance [Settles, 2009]. The AL process is represented in Figure 2.1., and it starts from getting a small amount of labeled data. After that, the model chooses only the most informative instances of the unlabeled pool of data to be labeled by an oracle or a single human annotator. The instances to which the model asks for labels are called queries. One can say that AL achieves its goal by allowing the machine to ask questions about the training data it possesses [Settles, 2011]. After the oracle labels the queried instances, they will be added to the training set, repeating this framework iteratively.

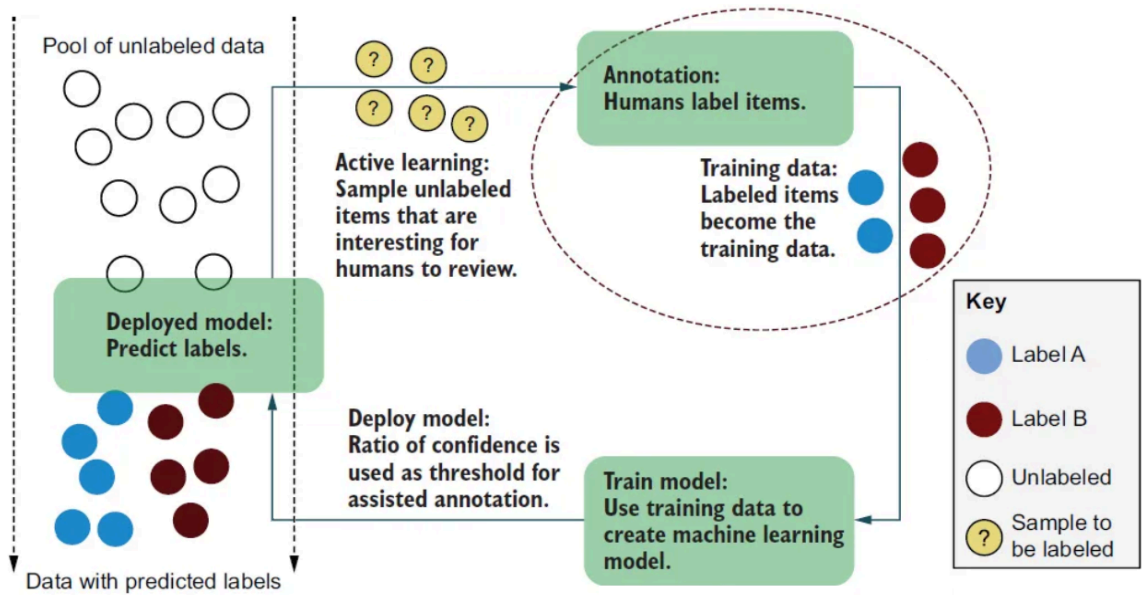

Figure 2.1 - Data annotation in Active Learning. Adapted from [Munro, 2021].

If random sampling captures the full diversity of data, AL is not needed [Munro, 2021]. The result is practically the same as labeling a random subset of data, which is faster and less expensive [Munro, 2021]. Unfortunately for ML workers, this is not often the case in most real-world scenarios. Thus, AL is usually recommended whenever only a small fraction of the unlabeled data can be annotated for limitations in time, money, or others.

### 2.2.1 Query Strategies

The way the process selects which data points to be labeled by the oracle is not random. The goal is to label the most informative instances for the model so that, with as few instances as possible, the model can know the information it needs about the data to achieve high accuracy faster. There are multiple ways to choose this data, known as Query Strategies.

**Uncertainty sampling**

One of the most commonly used query strategies is uncertainty sampling, proposed in 1994 [Lewis & Gale, 1994]. It entails strategies for identifying unlabeled data points close to the decision boundary of the current model, as presented in Figure 2.2. This means that the AL model requests labels for the instances it finds most ambiguous, which causes it uncertainty when trying to label them.

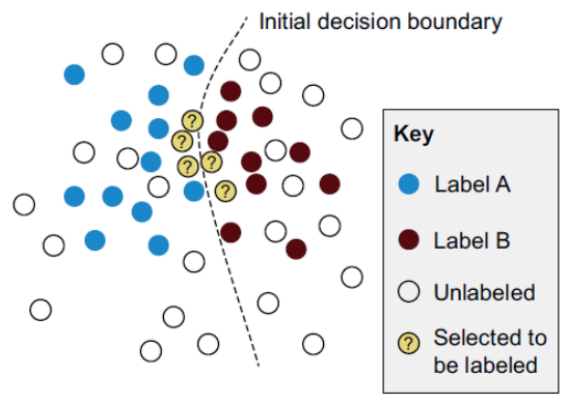

Figure 2.2 - One possible result from using Uncertainty Sampling. From [Munro, 2021].

The way we define uncertainty leads us to different approaches to this method [Munro, 2021, Settles, 2009]:

- Least Confidence sampling - Difference between the 100% confidence and the most confidence prediction. This is the most common approach. When we have a probability distribution over a set of labels $y$ for the instance $x$, the confidence can be quantified using the following formula, where $y^*$ represents the label with the highest predicted probability.

$$\varphi LC(x) = 1 - P(y^*|x)$$

We can now sample the instances where this value is the smallest, for those are the instances where the active learner is most unsure regarding the label.

- Margin of Confidence sampling - Difference between the two most likely predictions. It tells us how much more confident the model was in the label it predicted compared to its second-best option. We sample the instances where this difference is the smallest. The following formula defines this approach, with $y_1$ being the chosen label and $y_2$ being the second-best label prediction.

$$\varphi MC(x) = P(y_1^*|x) - P(y_2^*|x)$$

- Ratio of Confidence - Ratio between the two most likely predictions. This approach is similar to the last one, however, it may be viewed as a calculation of how many times the chosen label is more probable than the second-best label.

$$\varphi RC(x) = P(y_1^*|x) \, / \, P(y_2^*|x)$$

- Entropy-based sampling - Difference between all predictions, expressing how much each confidence differs from the others. This approach tells us how surprised we would be by each possible label relative to its probability. Instances with higher entropy indicate greater uncertainty about their predicted label.

$$\varphi ENT(x) = -\sum_y P(y|x)\log_2 P(y|x)$$

Figure 2.3 illustrates the target areas for sampling using three instances, according to different approaches. The heat of each pixel represents uncertainty. With this visual representation, it is noticeable how margin and radio of confidence focus mainly on pairwise confusion, while entropy maximizes confusion among all labels. The difference between the methods becomes pronounced when additional labels are introduced, highlighting the importance of consideration in selecting the appropriate strategy.

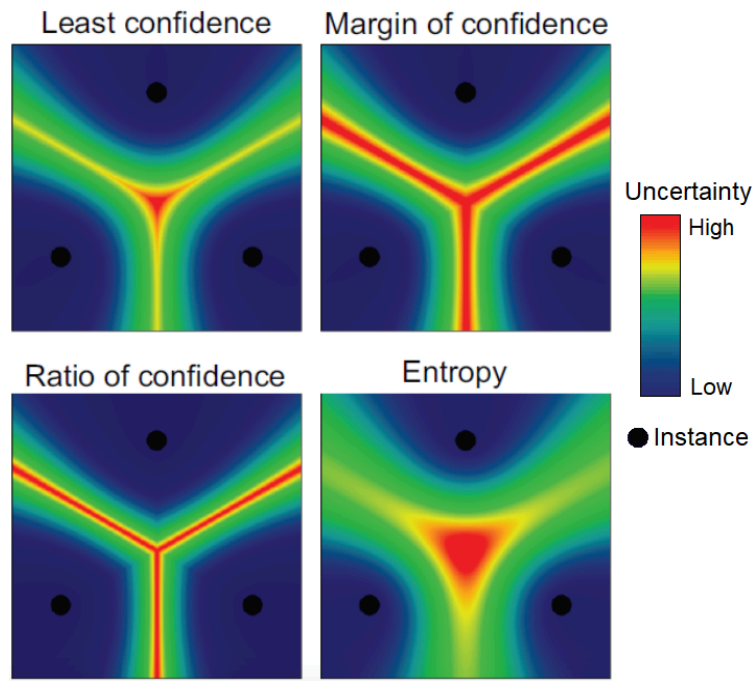

Figure 2.3 - Heat map of uncertainty sampling approaches and areas
that they sample for a three-label problem. Adapted from [Munro, 2021].

### Diversity sampling

Representative sampling, outlier detection, or stratified sampling, also known as diversity sampling, queries instances that are underrepresented or unknown to the current ML model. In practice, this strategy selects instances in different parts of the problem space. It is important to recognize that if diversity is situated far from the decision boundary, the selected instances are likely to be correctly labeled by the model, making human annotation potentially unnecessary. In Figure 2.4 we can see one possible example of how diversity sampling selects instances to be labeled by the oracle/annotator.

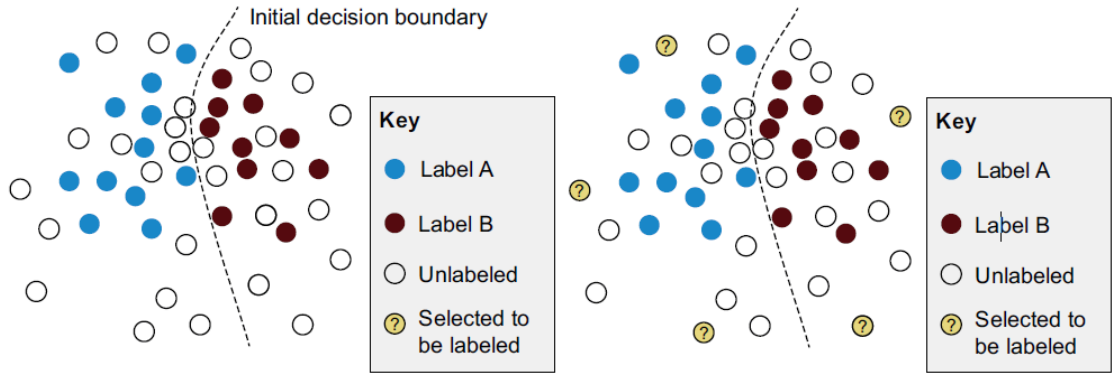

Figure 2.4 - One possible result from using diversity sampling.
Adapted from [Munro, 2021].

As in the previous section, we will explore the most common four approaches of this strategy [Munro, 2021] :

- Model-based outlier sampling - Identifies items that the model does not know yet. It is a statistical approach that identifies outliers while considering the underlying model of the data.
- Cluster-based sampling - Creates clusters that divide the data, helping to target a diverse selection, by making sure data from all clusters is used. It is the most used approach of diversity sampling in real-life problems due to its intuitiveness. Those familiar with clustering algorithms in UL, such as k-means, quickly understand its framework. It is a reasonable first choice when applying diversity sampling. Rather than focussing on studying the clusters, AL uses them to select instances for annotation by the oracle. In Figure 2.5 we can see one example of a clustering algorithm applied on data.

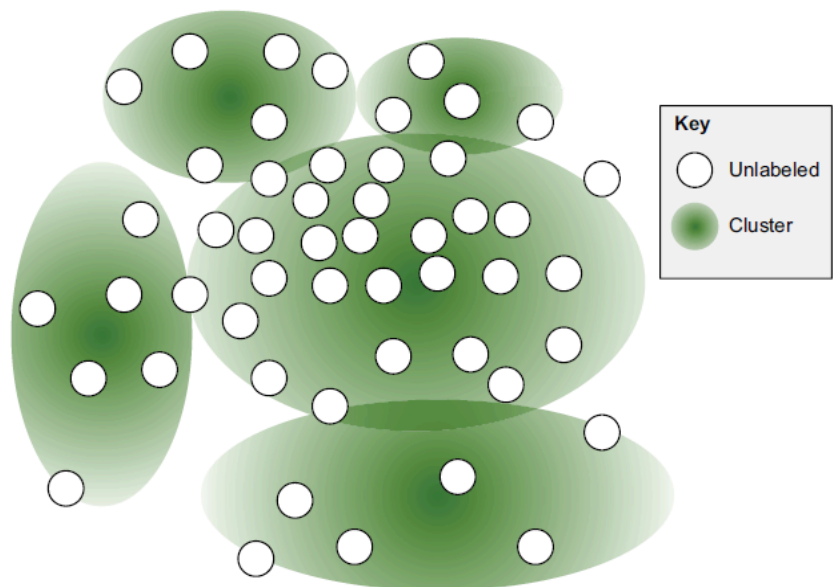

Figure 2.5 - One example of a clustering algorithm applied
in unlabeled data, resulting in five clusters. Adapted from [Munro, 2021].

Cluster-based sampling not only increases diversity, but also saves time. The oracle will annotate selected examples from each cluster, instead of multiple instances from the same one. This mitigates the risk of redundancy (common with random sampling in an unbalanced dataset) and ensures more heterogeneity.

- Representative sampling - Finds unlabeled samples that closely resemble the target domain in which the model will be deployed, as shown in Figure 2.6. It is one of the most powerful AL strategies [Munro, 2021], but it is also particularly susceptible to errors, overfitting, and noise, suggesting caution when used. For example, the unlabeled data might be very noisy, while the original given labeled data is clear. In this context, this sampling strategy would select instances that statistically resemble the target domain, but don't contain meaningful information for the task. This results in a lack of meaningful samples for labeling as a consequence of misleading the model.

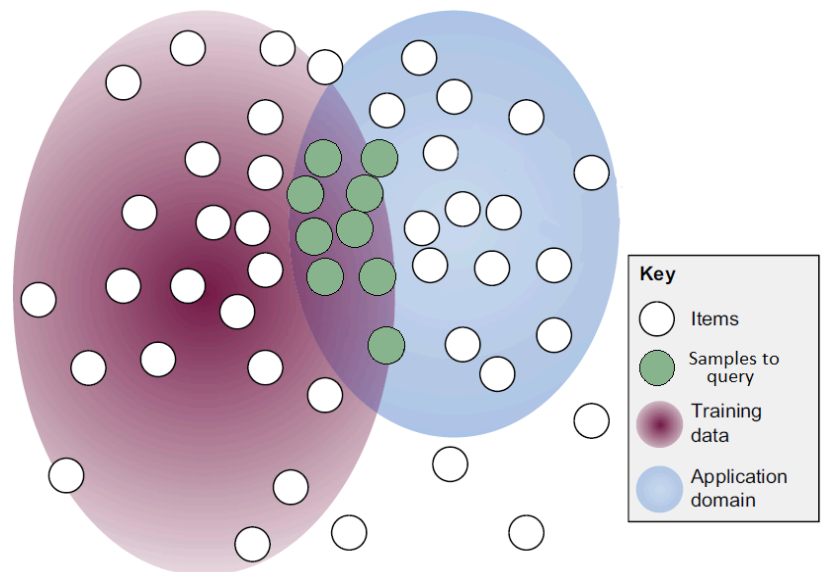

Figure 2.6 - An example of representative sampling. Adapted from [Munro, 2021].

Due to these limitations, Representative Sampling is not commonly used in isolation. One option is to combine it with uncertainty sampling, applying it only near the decision boundary, for instance.

- Sampling for real-world diversity - Ensures a diverse range of real-world entities in the training data to reduce bias. This strategy is particularly notable to consider when addressing factors like language, geography, gender, race, date, socioeconomics, or other demographics. The problem with bias and fairness in the model often arises from representativeness issues in the training data when certain demographics are over or underrepresented.

- Combined sampling - Combines uncertainty and diversity sampling in multiple ways to increase effectiveness. One combination (that is quite ideal for AL querying) is to sample instances that are near the decision boundary but the distance from each other, as in Figure 2.7.. Different combinations of these methods should be explored to find the most suitable for each specific task.

Even if only uncertainty sampling methods are implemented, inserting random samples for query is beneficial and easily applied. It is the most basic diversity sampling since every instance has the same probability of being queried.

**Query by Committee**

The most complex strategy for sampling in AL is Query-by-Committee (QBC) [Seung et al., 1992], where multiple models that are traditionally trained with the same labeled data, but still different (in parameters, for instance), share their opinions on labels. The instances where the models disagree the most will be selected as queries. In essence, disagreement between models in the committee is used as the informativeness measure in this approach.

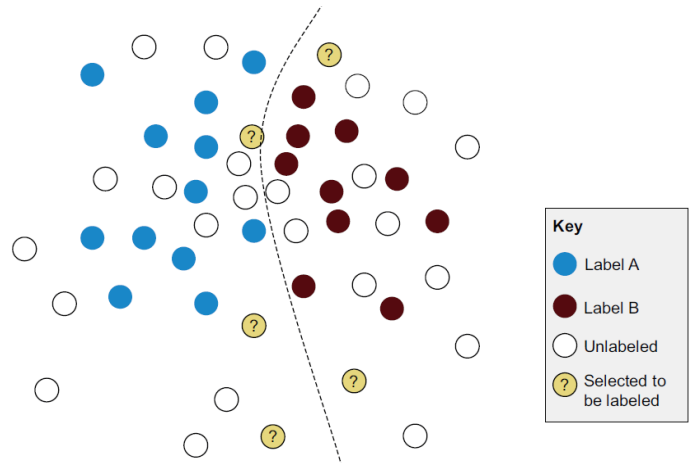

Figure 2.7 - One possible result from combining
uncertainty and diversity sampling. From [Munro, 2021].

QBC's goal is to minimize the version space, which represents all models/hypotheses aligned with the labeled data the model has until that moment. In Figure 2.8 there are two examples of version spaces shown, where the lines/boxes represent each model. Choosing the most informative instances to discard as many inconsistent hypotheses as possible reduces the version space. Thus, with less labeled data, the search for the best model in the version space in the target ML task is faster and more precise.

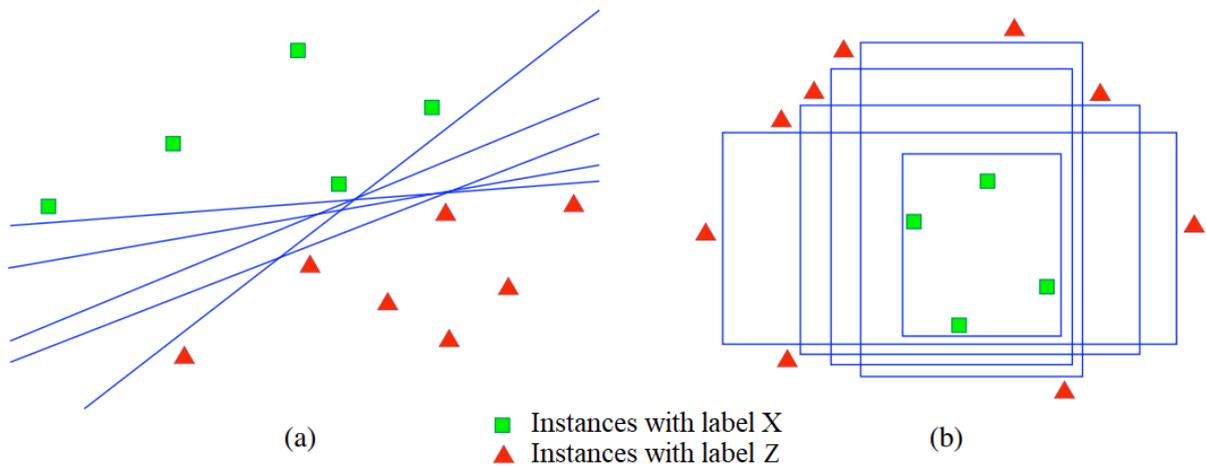

(a)    ■ Instances with label X
      ▲ Instances with label Z    (b)

Figure 2.8 - Version space examples for (a) linear and
(b) axis-parallel box classifiers. Adapted from [Settles, 2009].

To perform this strategy, one needs to assemble a committee of models that represent different areas of the version space, and to have some measure of disagreement between these models. Generative models form a committee by randomly sampling multiple hypotheses from a posterior distribution, each representing a plausible explanation of the labeled data. In contrast, discriminative models only need two or three ensemble methods, such as boosting and bagging, to produce a small but diverse set of models for the committee. Common approaches for measuring the level of disagreement are vote entropy [Dagan & Engelson, 1995] and Kullback-Leibler divergence [McCallum & Nigam, 1998].

### 2.2.2 Active Learning Sampling Scenarios

The active learner might select queries in different scenarios, and the three main settings considered in the literature are the following [Settles, 2009]:

- Membership Query Synthesis - In this setting the learner can generate samples for the oracle to label, reducing dependency on its unlabeled data. Although these synthesized queries can be suitable for various tasks, they may often be abstract and lack natural semantic meaning, being incomprehensible for individuals. As a result, if the oracle consists solely of humans, labeling can be challenging. Furthermore, these queries can be pointless in some fields. For instance, if these queries do not follow natural language rules in a natural language processing task;
- Stream-based Selective sampling or Sequential AL - In this scenario, the learner decides if each instance should be queried or not. It is especially useful when the acquisition of unlabeled data is low-cost or free. It also reduces annotation efforts compared to the pool-based AL (described next);

  There are multiple ways for the learner to decide whether or not to query an instance in this scenario:

  - Evaluating instances with informativeness measures - More informative instances are more likely to be queried [Dagan & Engelson, 1995]. This approach uses practices of both uncertainty and diversity sampling;
  - Computing region of uncertainty - Instances falling within the region of ambiguity of the learner will be queried [Cohn et al., 1994]. Similar to uncertainty sampling;
  - Defining a region that is still unknown to the version space - Instances that bring disagreement between two models of the same class, but with different parameter settings, will be queried. The main drawback is that calculating this region completely is computationally expensive and it should be updated after each query. This method is related to the QBC approach.
- Pool-based AL [Lewis & Gale, 1994] - In this approach, a large pool of unlabeled instances is considered, and the most informative ones will be queried. It is useful when large collections of unlabeled data can be collected simultaneously.

  Pool-based AL and Sequential AL contrast in the way they ask for queries. The former processes and ranks the entire unlabeled pool of data before selecting the best instance to query. On the other hand, the latter decides sequentially. For each query, the learner decides if it should ask the annotators to label it, or not. The Pool-based scenario is more commonly used in application papers, but it is generally less suitable when there is limited memory processing power.

### 2.2.3 Active Learning Annotation

In an ideal scenario, we would hope that the oracle is always right, inexpensive, and fast, however, realistically, this expectation is not met. Oracles can make mistakes, they need to be paid for their work, and the annotation process takes time.

In this section, we will introduce the topics of annotator selection, annotation quality control, and interfaces for annotation, as these are important fields that influence the whole annotation process [Munro, 2021].

**Annotators' Selection**

Misclassifications can occur when the annotation process is done by humans, creating unreliable annotations. Instances with incorrect labels become part of the active learner's training data, propagating errors and possibly introducing biases, as the model may overfit to those inaccuracies. Trying to reverse these problems, which ultimately lead to bad accuracy, is costly. Besides, these errors fail AL's objective of reducing costs while still achieving higher accuracy.

To reduce misclassification errors, annotators should be knowledgeable in the field of the task, but other factors influence their performance [Munro, 2021]. All humans are prone to mistakes due to cognitive factors, such as their focus and attention span, emotions, and decision fatigue, as well as contextual effects that can change their behavior. One example of a cognitive human factor is priming, which happens when a context or sequence of events influences human perception, affecting performance [Munro, 2021].

To minimize the effects of these influences on human perception, causing misclassification costs, it is crucial to identify skilled annotators who are also resilient to those factors. Furthermore, we may seek to mitigate the probability of these factors occurring, a topic covered later in the subsection about interfaces. We will now go over some of the most well-known factors in AL literature that influence annotators' performance and overall work [Herde et al., 2021a, Munro, 2021].

- Domain knowledge [Kazai et al., 2012: Munro 2021] - The skill and knowledge of an annotator have a vital impact on their performance, as mentioned before;
- Query difficulty [Beigman Klebanov & Beigman, 2010; Wallace et al., 2011; Whitehill et al., 2009] - The content or overall type of a query can make it more difficult than others. This can lead to a higher likelihood of errors in annotations;
- Ability for reliable self-assess - The design of the query interface can allow annotators to provide their confidence scores along with the label. If used honestly and correctly, this would give more reliability to labels. Despite empirical studies showing that in some domains annotators are accurate in this assessment [Calma et al., 2018a; Wallace et al., 2011], this is not always the case. The Dunning-Kruger effect states that unskilled annotators make more misclassifications and either do not realize their mistakes, or they are not honest about their confidence scores [Gadiraju et al., 2017; Kruger & Dunning, 2000];
- Motivation or level of interest - The engagement of annotators can influence the thoroughness applied during the annotation process. A study concerning crowdsourcing [Kazai et al., 2012] concluded that interested annotators performed better;
- Payment - The fair and regular compensation provided to annotators, whether they are in-house subject-matter experts or outsourced workers annotating for a short time, plays a role in their performance [Kazai et al., 2012; Munro, 2021];
- Ownership - The feeling that work done matters is crucial for annotators' motivation, fulfillment, and quality work [Munro, 2021]. Each worker should feel as much ownership as any other who worked for the same amount of time. For that, companies need to be transparent regarding the impact of their work and their daily or long-term goals;
- Concentration and emotional state - The lack of concentration can cause mindfulness and fatigue, which affect performance [Caldwell et al., 2019; Calma et al., 2016], while good mood is related to more productivity [Miner and Blomb, 2010; Oswald et al., 2015]. More information on these factors is presented later in Chapter 3;

- Machine-Human interaction - The dynamics of how annotators interact with machines during the annotation process directly impact their performance, as it will be further discussed in the subsection on interfaces;
- Collaboration - The opportunity for annotators to communicate and collaborate improves annotation quality. Chang and co-authors [Chang et al., 2017] show that presenting crowd workers with conflicting judgments and their justifications achieves higher accuracy than conditions with no collaboration.

All these factors must be taken into consideration to ensure the oracle is working at its best state. They can also be used as modifiable parameters when results are not as good as expected. For instance, one can try to pay workers more, give them break times to avoid fatigue or allow some form of collaboration between annotators as strategies to enhance performance.

**Annotation Quality Control**

In most real-world applications, majority voting decides which label will be used after multiple annotators provide their assessment of the queried instance. While the simplicity of this approach is appealing, it cannot be applied to all kinds of tasks. For instance, it is hard to define the right threshold for majority agreement if all annotators have seen different combinations of instances. Other and more trustworthy forms of annotation quality measurement must be used, as we must be confident in which annotations we consider reliable, and how we will select only one of the proposed labels as the correct one. Here are methods for measuring annotation quality [Munro, 2021] :

- Comparing with ground truth answers - This method is one of the simplest and most powerful for measuring annotation quality. Ground truth answers are a set of verified answers that are used to test annotators' accuracy and, therefore, reliability. If an annotator correctly labels 80% of these instances, one can say that they are 80% accurate. The probability of any given label being incorrect is determinable knowing the reliability of each annotator and the number of workers that labeled each instance. Ground truth data can also be used to establish baselines for expected accuracy (adjusted for chance) by evaluating the performance of annotators against ground truth data. This aids the decision of which annotations are reliable and which should be discarded.

  To guarantee that ground truth answers are correct, we must use combinations of the other methods listed below. Regarding how to select instances as ground truths, one might get random samples from the whole data or from the current iteration to ensure they are from the same distribution, or samples selected deliberately for their usefulness in annotation guidelines.

- Inter-annotator agreement - This metric can be used to learn a lot about each annotator and instances. Although model accuracy and agreement among annotators should not be compared directly, the latter might help us make decisions about the quality of annotations and annotators themselves:
  - If inter-annotator agreement is consistently high, we can have more trust in the reliability of the model;
  - If an annotator frequently agrees with others, they are likely the most accurate, while those who often disagree may be confused or unreliable;
  - If an annotator shows inconsistency in performing the same task at different times (intra-annotator disagreement) it may indicate a lack of focus or too much subjectivity in the query. In this case, one should inquire about the cause of this inconsistency. If a change of opinion was not the motive, this occurrence can be used as an indication to change the format of the queries;

○ If inter-annotator agreement decreases with a new data source, we can expect the model's accuracy to decrease as well. As so, this metric may be used to evaluate the inherent difficulty of an ML problem;

○ If annotators frequently disagree, the task might be very subjective, allowing for multiple valid annotations. In this case, disagreements are often beneficial, as they suggest diversity among the annotators, thereby reducing the likelihood of demographic bias in the annotations;

○ If a task has a low agreement between less-qualified workers, one might consider routing it automatically to an expert.

Inter-annotator agreement should be used combined with ground truth data to extend accuracy analysis [Munro, 2021]. For example, annotators' accuracy assessed against ground truth data can serve as confidence measures when combining multiple annotators for a task. Then, it is easier and faster to analyze each group's intra-agreement since the overall reliability is known.

● Expert review - This approach is one of the frequently adopted methods of quality control. Some instances should promptly be given to experts, who are rarer and more expensive than other workers. Their annotations can be used as ground truth or as a decision-making technique for instances with low inter-annotator agreement. They might even be used to create guidelines and training material in the form of examples, as illustrated in Figure 2.9.

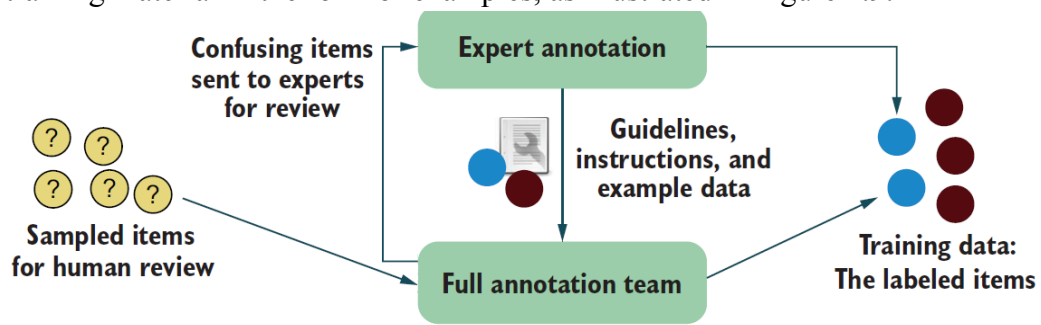

Figure 2.9 - Workflow for experts incorporated in
the annotation process. Adapted from [Munro, 2021].

● Multistep workflows and review tasks - This approach decomposes a challenging task into simpler, smaller subtasks, thereby simplifying the process of quality control. As this technique promotes singular focus it increases concentration, accuracy, and speed when annotating. The main drawback is the added difficulty in managing these more elaborate workflows. Figure 2.10 portrays one example of this approach, which also includes a review step, used to mitigate errors, promoting higher accuracy and reliability.

**Interfaces for annotation**

The design of the interface used for the machine-human interaction in annotation can mitigate some of the factors that affect workers' performance. In general, a good interface should have the following elements [Munro, 2021] :

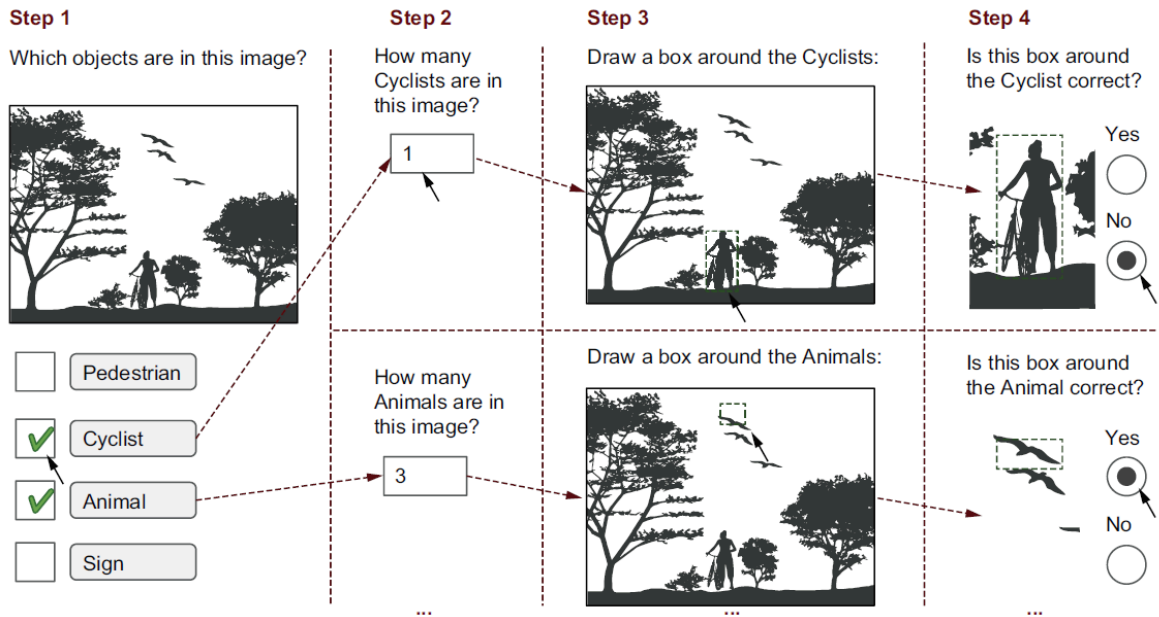

Figure 2.10 - Example of a multistep workflow on an object labeling task. From [Munro, 2021].

- Affordance - Objects should function as people expect them to. This reduces confusion ensuring a more intuitive use of the platform;
- Feedback - Acknowledgment of user actions must be apparent, whether through a message or an animation. Feedback validates affordance, indicating clearly whether the expectations regarding the functionality of the interface are met;
- Agency - Annotators should feel ownership and a sense of power in their actions. Affordance and feedback assist this feeling.

There is also more general advice for how to structure and how to use the interface [Munro, 2021] :

- Minimizing eye movement - The placement of all elements needed for annotation should remain consistent regardless of the data's varying attributes. This consistency is crucial since eye movement compels the annotator to divert their attention across the screen to find what they need, reducing efficiency;
- Minimizing scrolling - The requirement for scrolling should be eliminated, as it leads to fatigue and wastes time. When content that could fit on the screen is only visible after scrolling, users become less attentive and more frustrated;
- Fitting language direction bias - The interface should be biased toward a left-to-right layout if the annotators' language employs a left-to-right writing system, as it becomes more intuitive;
- Allow keyboard shortcuts - The use of keyboard shortcuts can increase the speed of input and navigation. The sequence of actions triggered by them should feel natural to the user, and the concept of affordance (mentioned above) should be present. Other tools can also be used with the same idea of efficiency, such as pedals to move audio or video records back and forward in time;
- Using ranking instead of absolute values when possible - The use of continuous absolute values is often unreliable, as individuals often have varying standards over time that also differ from one person to another [Thurstone, 1927]. Rankings, on the other hand, are a more dependable choice, since humans are better with simpler comparisons between two items.

In addition to optimizing the interface for efficiency directly, it is important to consider how it may impact factors indirectly too, such as users' perception and mood. For instance, interfaces that annotators enjoy working with promote better performance [Cakmak et al., 2010]. Conversely, interfaces can be prone to priming, a phenomenon in which the surrounding context or sequence of events influences human perception. This can promote errors that would not happen otherwise.

There are various studies regarding priming. Hay and Drager [Hay & Drager, 2010] is a well-known example where the task was to distinguish New Zealand and Australian accents. Either a stuffed kiwi bird or kangaroo toy (known animals from those countries) was placed on a shelf where the study took place, but they weren't mentioned to the participants. In the end, people classified more instances as the country correspondent with the toy present in their room, despite being wrong. The earliest documented research on priming was presented by Meyer and Schvaneveldt [Meyer & Schvaneveldt, 1971], which concluded, as similar to subsequent studies, that various factors can influence accuracy in perception tasks [Hay et al., 2006].

Repetition priming [Munro, 2021] is common on subjective tasks, where annotators may change their opinion based on the items they have seen before. This typically occurs when there is a change in the boundary separating neighboring categories, such as distinguishing a negative or very negative sentiment on a sentiment analysis task. Furthermore, repetition in the labels given by the annotators increases the likelihood of the annotator experiencing fatigue, distraction, and frustration [Amershi et al., 2014]. Consequently, when an instance with another label appears, annotators are more inclined to mindlessly misclassify. Randomizing the order of instances to label is easy to implement and may help in avoiding consecutive items being similar. However, if the dataset is imbalanced, a diversity sampling method (like cluster-based), should be implemented instead.

The interface should also have a mechanism that allows annotators to give feedback, in free text form, about any aspect of their tasks [Munro, 2021]. This can include comments on their knowledge limitations, clarity of instructions, intuitiveness of the interface, or patterns noticed in the data. Understanding the perspectives of annotators is advantageous for everyone involved, as it helps maintain a shared understanding.

It may also be beneficial to elicit what annotators think their co-workers would annotate, as it improves objectivity [Munro, 2021]. This happens because the annotator distances themselves from their own perspective, reducing personal biases and emotions, and giving an answer that is closer to their true objective opinion. Bayesian Truth Serum [Prelec, 2004] is a metric that uses this idea as its base. This method also allows annotators to be more honest and more comfortable annotating a valid minority opinion. It also makes it easier to convey negative responses that workers might be reluctant to convey for various reasons [Munro, 2021].

## 2.3 Recommendation Systems

Recommendation Systems (RS) are software tools and techniques that suggest items or content to users based on their preferences, interests, the behavior of similar users, or items' contextual information [Raza et al., 2024]. Item profiles have information describing them, such as genre, type, year of release, and age rating, while user profiles include demographics, past ratings, and browning history.

Web applications frequently implement RS to improve user engagement by suggesting content based on their (or similar users') interactive behaviors. RS's main goal is to help users make decisions and improve their experience using the application. It helps users discover content that is more relevant to them [Raza et al., 2024].

RS is especially useful in Big Data contexts, as it effectively identifies recommendations amidst vast quantities of user interactions and items to recommend. Additionally, various fields use RS to achieve different kinds of goals, such as personalized education (adapting learning strategies to different students) and healthcare (assisting in the choice of the most appropriate treatment for a patient) [Raza et al., 2024]. Combinations of Large Language Models with RS have also been developed, enabling the systems to process text data, enhancing their abilities by increasing accuracy and having more dynamic recommendations [Raza et al., 2024].

Traditional RS methods can be divided into three main categories: Collaborative Filtering (CF), Content-Based Filtering (CBF) (which includes Knowledge-Based RSs), and Hybrid approaches.

**Collaborative Filtering**

This approach assumes that similar users have similar tastes. Therefore, if one user enjoys a particular item, it is likely that another user with similar preferences will feel the same way [Raza et al., 2024]. To do this, CF exploits information about a similar user's past interactivity (namely, watch time and reviews) and uses it to predict what the current user would be interested in [Jannach et al., 2010].

Figure 2.11 presents a simplified example of CF. Checkmarks represent a high rating given to a book by the respective user, while the "X" symbol means a bad rating. To decide if "Book 3" is a good recommendation for the user who has not read it, the RS will use the user's most similar to them to predict. In this case, the 3rd user is the most similar, and they enjoyed "Book 3". Therefore, the RS will recommend this book.

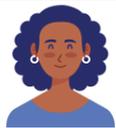

Figure 2.11 - Representation of a Collaborative-
-Filtering RS. Icons from flaticon.com.

A traditional pure CF approach uses a matrix of user-item ratings as input to produce either a list with *n* recommendations (unknown to the current user), or a numerical prediction that represents how much the individual will appreciate a possible recommendation [Raza et al., 2024].

CF is the most used technique in commercial RS, for instance, online retail sites. These websites use CF to identify similar buyers, which helps suggest additional products that a given user is likely to appreciate, ultimately increasing sales [Jannach et al., 2010].

This approach does not need to know anything about the item, which makes it easier to apply, and avoids the costly task of updating their descriptions whenever changes occur. However, its main drawbacks are scalability, sparsity, and cold-starts [Jannach et al., 2010; Raza et al., 2024; Sharma and Mann, 2013].

The cold-start problem is a common problem in RSs, where the system can not make good recommendations due to the absence or scarcity of data about the user or item [Jannach et al., 2010; Pozo, 2018]. For this reason, this approach may not be effective for items that are rarely purchased, such as houses or computers, due to the scarcity of available information on purchases.

**Content-based Filtering**

This method relies on the user's past interactivity and item' characteristics to generate recommendations [Ibrahim and Saidu, 2020, Sharma and Mann, 2013]. Their profile consists of a vector that assigns different weights to various item attributes, reflecting the significance of each attribute to the user [Ibrahim and Saidu, 2020]. The main task is to identify items that are most aligned with each user's individual preferences. CBF relies on products' textual descriptions and features. It may use textual descriptions and features of products to form recommendations using information retrieval techniques such as Term Frequency-Inverse Document Frequency. This allows it to assess the relevance of item attributes which, when used with similarity measures like cosine similarity, can be compared to make recommendations. ML techniques, including Neural Networks and clustering algorithms, can also be employed to enhance the recommendation process by learning patterns from the underlying data [Sharma and Mann, 2013].

In Figure 2.12 a representation of a CBF system is presented. In this representation, the colored bars beneath each book signify various genres, and the checkmarks show the user's enjoyment of the book underneath, while the cross mark represents their dislike. The RS suggestion is made, represented with a blue question mark, based on the user's ratings of books and their genres. In this case, the RS recommends a book with genres that the user enjoyed before, and that does not contain genres that they disliked in the past.

A good advantage of this approach is that there is no need for a large user community with a rating history to create suggestions, avoiding the cold-start problem. Recommendations can be given even if there is only one user, since it only uses the information of the person who is being recommended an item [Jannach et al., 2010].

One drawback of this approach is that, although technical descriptions of items' characteristics are provided by the manufacturers (such as the genre of a book, actors in a movie, or singers in a song), qualitative features are hard to obtain. This happens because these values are frequently subjective. For example, the taste or quality of a product might have to meet different conditions (that also vary with personal assessment) to be classified as a certain category. Additionally, CBF can also struggle to recommend unseen or new items [Raza et al., 2024].

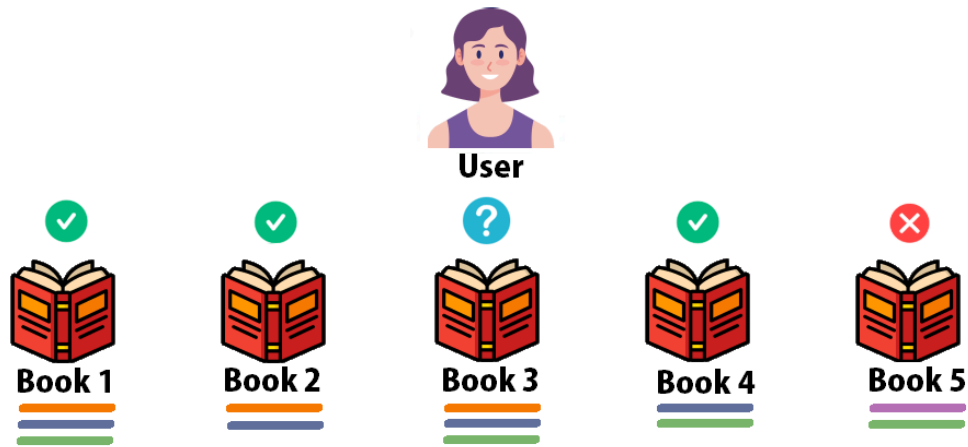

Figure 2.12 - Representation of a Content-based RS. Icons from flaticon.com.

One variant of CBF is Knowledge-Based RS [Jannach et al., 2010]. This alternative is particularly relevant when there is very limited historical data regarding the relationship between users and items. The application fields tend to be around items that are seldom bought, such as houses, cameras or computers. In these cases, since there is a low amount of ratings from the community, a pure CF RS will not be a good fit.

In Knowledge-based RSs, recommendations are determined separately from the ratings given by individual users: either by assessing the similarities between customer needs and products, or by applying explicit recommendation rules. In these systems, there is no specific need for previous ratings from users [Jannach et al., 2010]. By giving information about what we are looking for, the system recommends an item based on that interaction. This interactivity shows how RSs can be seen as a guide to find personalized interests from a large space of options, instead of the traditional use as a personalized filtering system [Jannach et al., 2010].

There are two basic types of Knowledge-based RS: constraint-based systems and case-based systems. In constraint-based systems, the RS checks which items the set of recommended items fulfill the recommendation rules, recommending only those. In the latter, items are retrieved by using similarity metrics between them and the expressed requirements [Jannach et al., 2010].

Moreover, given the rapid evolution of some fields, such as technology, outdated reviews may become irrelevant, making a standard CBF an unsuitable choice as well. Knowledge-Based RS solves these problems by creating knowledge based on the information it has about both the user and the available items. Additionally, this type of RS allows users to formulate their requirements explicitly, taking into account specific information important for the user's preferences. This is an advantage not provided by either CF or pure CBF.

**Hybrid RS**

This approach combines the strengths of CF, CBF, and other existing approaches such as Knowledge-based filtering or Demographic-based filtering (based on user demographics). It uses different recommender components to offer better recommendations. It is used to overcome some of the singular approaches' disadvantages. This approach can be done in multiple ways, combining multiple recommender techniques and also other ML components (such as Neural Networks or Bayesian Networks).

Figure 2.13 illustrates an example of a monolithic design Hybrid RS. It is a recommendation component that integrates various recommendation strategies by preprocessing and merging various knowledge sources.

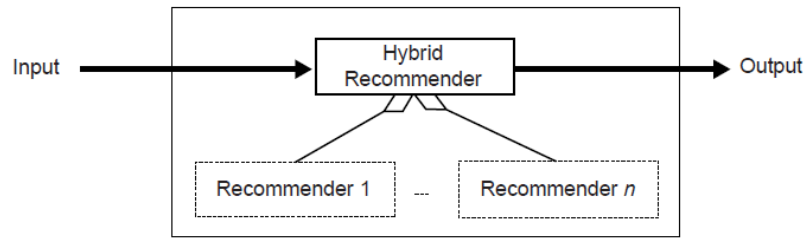

Figure 2.13 - An example of a Hybrid RS - a monolithic hybridization design. From [Jannach et al., 2010].

Figure 2.14 presents another use of hybrid RS, where multiple RS operate independently and generate separate recommendation lists.

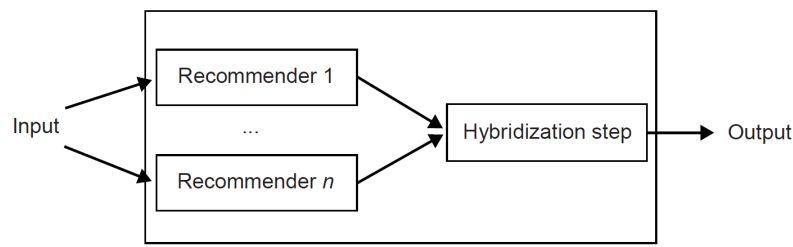

Figure 2.14 - An example of a Hybrid RS - a parallelized hybridization design. From [Jannach et al., 2010].

## 2.4 Chronotypes

Every living creature that has a life span of more than several days has an inherent biological cycle known as the circadian rhythm [Walker, 2017]. This cycle controls all physiology levels, such as preferred times for eating and sleeping, body temperature, mood, emotions and productivity, by sending signals to the organs. In humans, this cycle lasts twenty-four hours, due to the synchronization with the solar cycle [Clark et al., 1989; Roenneberg et al., 2019; Sládek et al., 2020; Walker, 2017].

The protein components involved in each person's circadian rhythm may vary as a result of genetic variance, leading to the variability in its timing (being earlier or later) [Roenneberg et al., 2019]. In practical terms, individuals may be genetically predisposed to sleep, wake up, eat, be productive, or exercise at different hours of the day. These inter-individual differences of synchronization with the solar cycle are called chronotypes, and its variance in the population tends to follow a normal distribution [Kalmbach et al., 2017; Roenneberg et al., 2019; Sládek et al., 2020].

These differences in chronotype colloquially classify people as larks, hummingbirds or owls [Breus, 2016; Roenneberg et al., 2019] to designate early, regular or late risers, respectively. However, in his book, Breus [Breus, 2016] denotes concern about the Morningness-Eveningness Questionnaire used to classify individuals in one of these three categories. The writer states that, from his practical experience as a clinical psychologist and a sleep medicine expert, this test does not effectively classify chronotypes, but merely assesses preferences for wake-up times. To address this problem, the author created a new test, and a new classification system, dividing into four categories: Dolphins, Lions, Bears and Wolves.

- Dolphins: Individuals who have a hard time falling asleep and waking up. Much like real dolphins, their brain doesn't rest completely when sleeping, being light sleepers and waking with small noises. These people are more productive from 3 PM to 7 PM. They account for 10% of the population;
- Lions: People who are morning-driven, naturally waking up earlier and being more focused in the morning. These people are more productive from 9 AM to 2 PM. They account for 15-20% of the population;
- Bears: Humans who sleep better and whose cycle matches the sun the most. They are most productive when the sun is at its highest point. These people are more productive from 10 AM to 2 PM. They account for 50% of the population, being the most common chronotype;
- Wolves: Individuals who prefer working late. They usually start to feel sleepy when lion types feel the most productive, just like real wolves. These people are more productive from 1 PM to 5 PM. They account for 15 to 20% of the population.

The author [Breus, 2016] states that these range of chronotypes exist due to evolution, and that they can change during one's lifetime. The presence of diversity was essential to ensure that humans could maintain safe watch shifts during the night when sleeping in the wild. Given that the categories identified by the author exhibit similarities to the sleep patterns of other mammals, Breus [Breus, 2016] designated the chronotypes in reference to the animals that most closely resemble each of them.

Chronotypes exhibit significant variations between biological sexes, and are subject to continuous change over the course of an individual's lifetime, particularly in relation to age [Breus, 2016; Fischer et al., 2017; Kalmbach et al., 2017; Sládek et al., 2020]. Figure 2.15 shows the different chronotypes over age and gender using over fifty thousand respondents (approximately thirty thousand women and twenty three thousand men) in a scatter plot with a fitted polynomial curve and shading for confidence intervals (Figure 2.15a), and in a histogram (Figure 2.15b).

Another way to visually assess chronotypes is to use Best Alertness interval midpoint (BAmid) and Mid-Sleep on Free days corrected (MSFsc). MSFsc is the adjusted mid-sleep phase on non-working days, with sleep deficit accounted for, presented in hours relative to midnight.

Figure 2.16 shows the MSFsc and BAmid of different ages and sexes, as well as the means ± standard deviation, and the fitted polynomial curve (where shading indicates the confidence interval from bootstrap). The data is consistent with the overall conclusions from Figure 2.15, reflecting a normal distribution over the population, despite using a smaller number of respondents (over 3 thousand).

The data in Figure 2.15 and Figure 2.16 supports that, as individuals age, their chronotype tends to shift toward an earlier schedule. Furthermore, while women typically display an earlier chronotype than men until they reach the ages of 40 to 45, this changes. This difference becomes more noticeable after the age of 60.

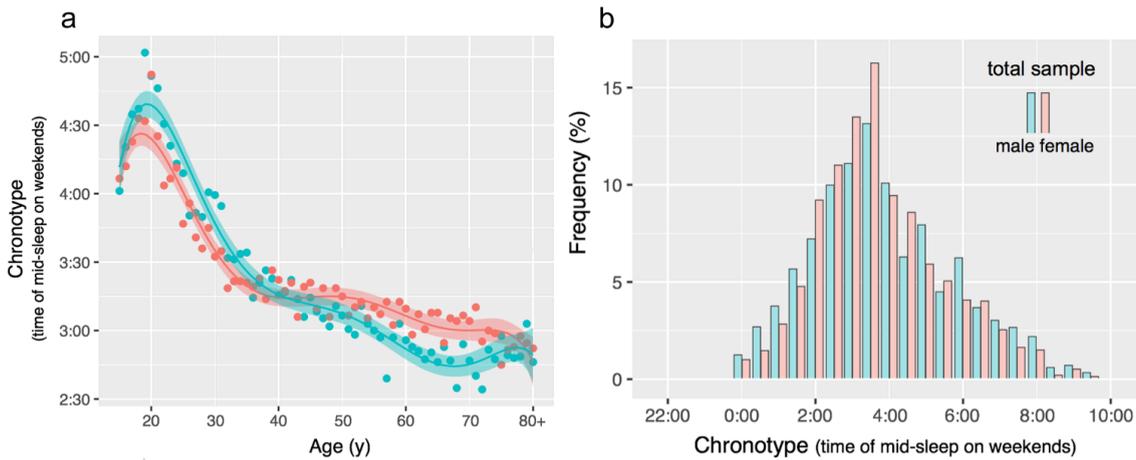

Figure 2.15 - a) Chronotype distributions over age and sex. b) Chronotype frequency of the overall respondents. Adapted from [Fischer et al., 2017].

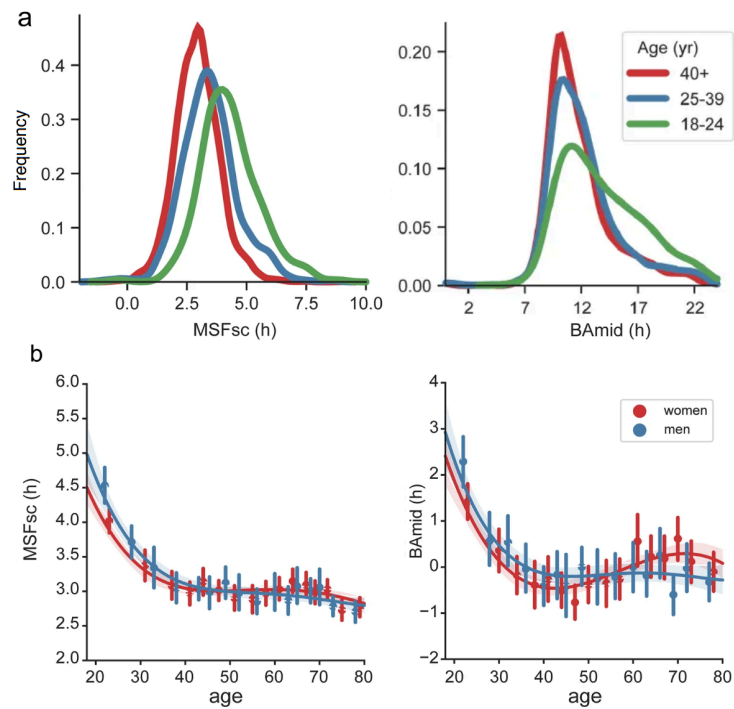

Figure 2.16 - a) Distribution curve for chronotype (MSFsc and BAmid) across different age groups categorized by age. b) Chronotype over age of respondents. Adapted from [Sládek et al., 2020].

Research has identified circadian variations in mood states, indicating significant correlation in wakefulness levels and mood [Kline et al., 2010; Koorengevel et al., 2003; Monk et al., 1997; Murray et al., 2002]. Other studies [Clark et al., 1989; Murray et al., 2002] analyze its correlation with mood by using Negative Affect (NA) and Positive Affect (PA) calculated with the PANAS scale [Watson et al., 1988], where both can vary between a score of 10 and 50. NA references a range of strong unpleasant emotions, such as fear, anger, and sadness, with a low value indicating calmness. On the other hand, PA reflects engagement and positive emotions, namely joy and enthusiasm, with a low PA indicating lack of enthusiasm.

Murray and co-authors [Murray et al., 2002] concluded that mood values (calculated using PA and NA) closely followed the values of energy through the day, caused by the circadian rhythm, as visible in Figure 2.17. The mood values exhibit a slight increase prior to wake-up time, followed by a gradual decline in the afternoon.

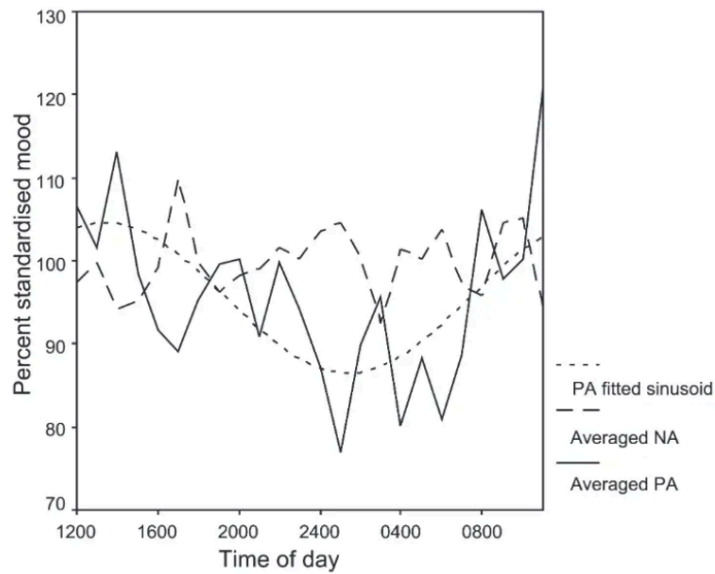

Figure 2.17 - Average PA and NA values, and a 24-hour
sinusoidal curve fitted to PA values. From [Murray et al., 2002].

Clark and colleagues [Clark et al., 1989] reached similar conclusions, succinctly presented in Figure 2.18. In college students, NA values are not affected by circadian rhythms, being very stable through the day. However, PA levels increased from 9 AM to noon, and decreased by a similar amount from 9 PM to midnight. This indicates similarity to the natural energy from circadian rhythm.

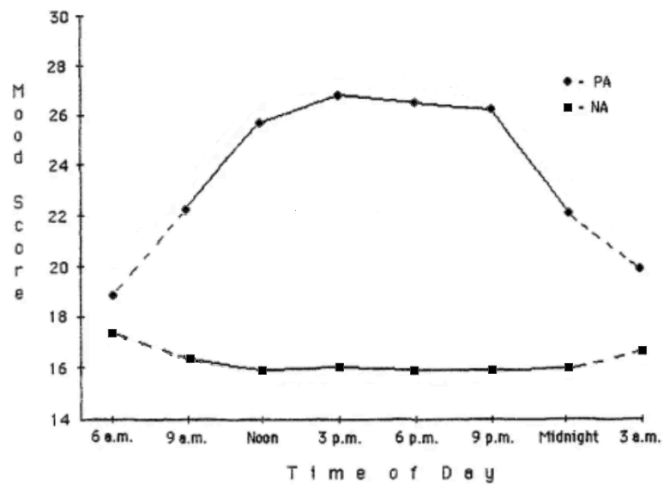

Figure 2.18 - Diurnal pattern of PA and NA values.
Adapted from [Clark et al., 1989].

Furthermore, the same authors concluded that individuals that do not identify themselves as morning or evening people, behave like morning types in all respects. In Figure 2.19, they also present the NA and PA results for self-labeled "morning" and "evening people". Morning people were higher in PA scores during the whole day, indicating that earlier chronotypes tend to have better mood and positive emotions than individuals who have a delayed cycle.

Monk and colleagues [Monk et al, 1997] concluded that circadian temperature rhythm serves as a reliable predictor of circadian performance rhythms. Additionally, the authors provided evidence that alertness, referred to as global vigor, is strongly correlated with reasoning accuracy. Figure 2.20 shows global vigor plotted against the hour of the day (with a linear decreasing trend removed). It illustrates how alertness follows a pattern similar to energy throughout the circadian rhythm.

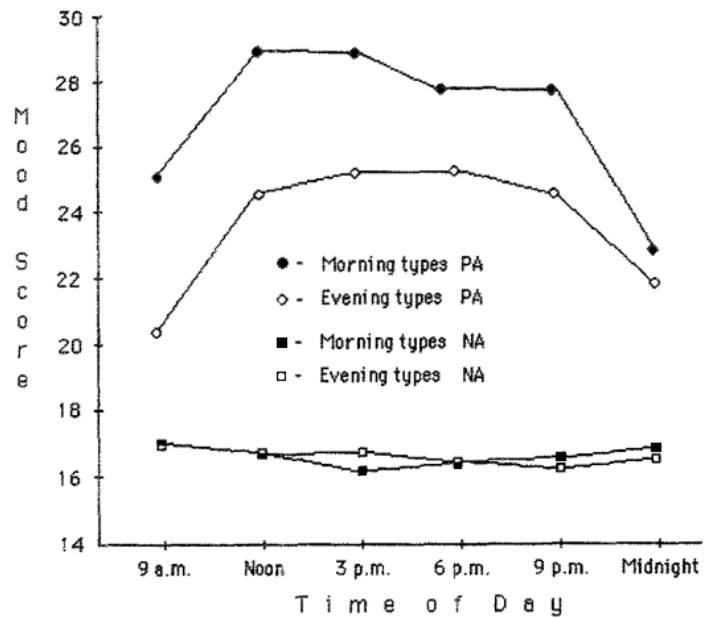

Figure 2.19 - Diurnal pattern of PA and NA values for morning and evening types. Adapted from [Clark et al., 1989]

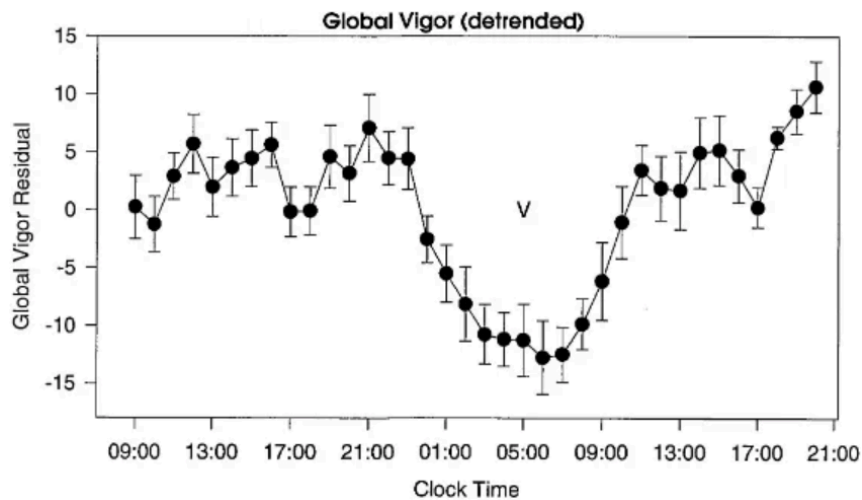

Figure 2.20 - Global vigor levels across hours of the day after removing a linear decreasing trend. From [Monk et al, 1997].

Since chronotype influences energy levels and wakefulness, and energy is linked to better focus and mood, chronotype also affects concentration and performance [Breus, 2016]. In Figure 2.21 McHill and colleagues [McHill et al., 2018] in their study deviate from a 24-hour day. The circadian phase is indicated in degrees, with zero degrees representing midnight. Despite the presence (black) or absence (white) of sleep deprivation and the disregard for the natural human circadian rhythm, lapses in attention and subjective alertness remain correlated with a conventional chronotype.

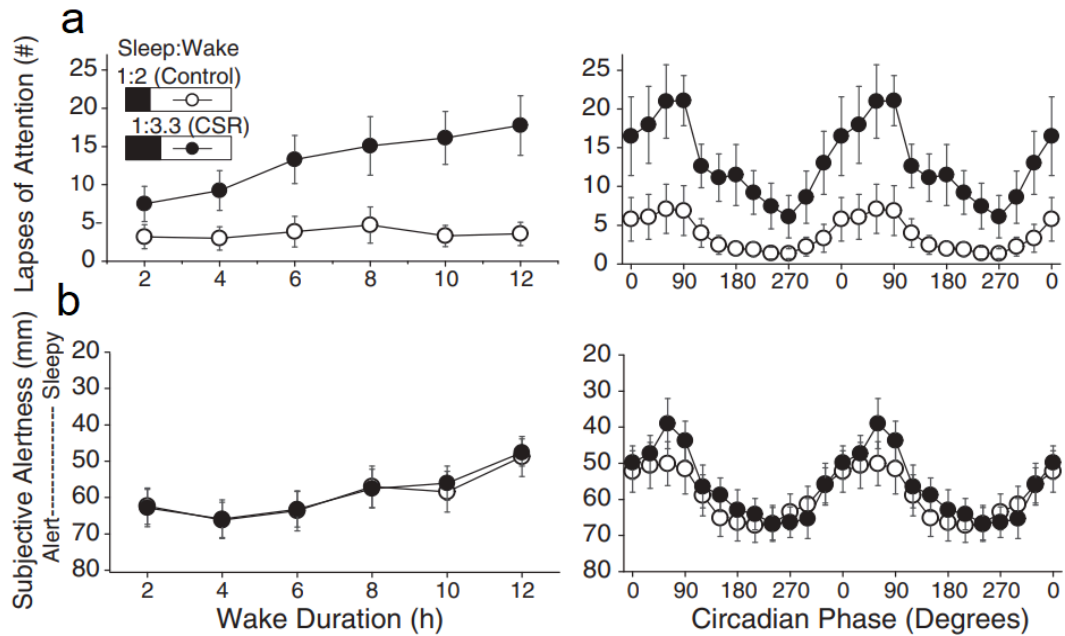

Figure 2.21 - Lapses of attention (a) and subjective alertness (b) plotted on wake
duration and circadian phase, shown for the control condition (white)
and chronic sleep restriction (black). Adapted from [McHill et al., 2018].

## 2.5 Chapter summary

This chapter delves into the essential topics pertinent to the work, starting with the
notion of AI and ML to introduce the original and context around AL. It focuses on
explaining what AL is, and how it is usually applied, including the Query Strategies
used to select which instances to send to the annotators, as well as presenting the most
common sampling scenarios.

The chapter further introduces concepts of imperfect annotators, presenting different
facts that influence their productivity at work, as well as ways to improve annotation
quality control. We also explore techniques on how to make interfaces used for
annotation less error-prone, in order to improve the quality of annotations.

Additionally, this chapter provides a brief overview on RSs, explaining their overall
goal and different types. We introduce here the origin of Knowledge-Based RSs, and
how they are traditionally used.

The chapter concludes by briefly introducing the concept of chronotypes. Research on
how they vary by age and sex is presented, focusing on their reported correlation with
mood and concentration.

# Chapter 3
# State of the art

This chapter reviews the analysis of existing approaches and solutions that directly or indirectly fall within the scope of this work. The sections that mention AL directly are heavily based on the organized survey by Herde and co-authors [Herde et al., 2021a].

The upcoming sections will start by discussing how different authors work around the existence of imperfect annotators. It mentions how emotions, fatigue, and the inability to show uncertainty may contribute to imperfect annotations and how various authors dealt with this influence. After that, the chapter focuses on how previous papers select the query-annotator pairs aiming to improve the speed or/and accuracy of their work. Then we explore how AL and RS have been used together in the past and we conclude the chapter with a discussion of the related works and their similarities and differences with the present work. Finally, we examine insights and implications of the literature presented in our study.

## 3.1 Imperfect Annotators

The existence of imperfect annotators is a well-known limitation in AL. Multiple factors influence an annotator's performance, as mentioned in Chapter 2. These factors can be internal (such as mood, motivation, and fatigue levels), or they can even be external (namely difficulty of the query and working environment). The values of these features either vary from the period in which the annotation is done, or from one query to another, which affect annotators' performance, increasing or decreasing the likelihood of making errors when annotating. As Munro [Munro, 2021] states, when the annotators used are human, they will always be imperfect when labeling. Therefore, the strategy used for selecting the most suitable annotators should take this into account to avoid misclassifications.

### 3.1.1 Influences on Annotation Performance

While there are multiple factors affecting one's productivity when labeling, most papers only consider three influences on the annotator's performance: the type of annotator, the type of query, and the optimal solution. This means that, usually, three types of performances are assumed [Herde et al., 2021a]:

- Uniform annotator performance: The annotator performance depends only on the characteristics of the annotator. Usually, those characteristics focus on the annotator's skill (for instance, their performance in the present or the past, or their specialty);
- Annotation-dependent annotator performance: The annotator performance depends on the annotator's characteristics and the optimal annotation for a query. For example, if the correct label is usually more challenging to get right;
- Query-dependent annotator performance: The annotator performance depends on the annotator's characteristics, the optimal annotation, and the query itself. For instance, consider the scenario where a specific query is usually mistaken for another label, which likely reduces the annotator's accuracy. It is important to take the additional difficulty of that label into account.

**Uniform annotator performance**

Regarding uniform annotator performance, Donmez, Zheng, and respective colleagues [Donmez et al., 2009; Zheng et al., 2010] use majority voting to select the true label of an instance. Then, using the interval estimation method [Kaelbling, 1993], the majority votes are then used to estimate the maximum proportion of instances that have been correctly labeled by each annotator. More recently, Donmez and co-authors [Donmez et al., 2018] considered the performance of each annotator changes over time. That change follows a Gaussian distribution with a zero mean and a defined variance, shared among all annotators. Long and associates [Long et al., 2013, 2016; Long and Hua, 2015] estimate an annotator's performance by comparing their labels to the estimated true annotations. This value indicates the probability that this annotator correctly labels an instance.

**Annotation-dependent annotator performance**

Regarding annotation-dependent annotator performance, Moon and Carbonell [Moon and Carbonell, 2014] select a label as correct with majority voting, and then measure each annotator's performance by calculating their accuracy in labeling each class correctly. Rodrigues and colleagues [Rodrigues et al., 2014] use Gaussian processes and expectation propagation to assess how accurate each annotator is. It compares their annotations to the label chosen as correct, measuring both sensitivity and specificity for different classes. Nguyen and co-authors [Nguyen et al., 2015] estimated crowd workers' performance by comparing their labels to the ones given by experts.

**Query-dependent annotator performance**

Regarding query-dependent annotator performance, Wallace and colleagues [Wallace et al., 2011] assume that higher pay indicates better performance. Donmez and associates [Donmez et al., 2008; Donmez and Carbonell, 2010] use annotators' confidence scores to estimate their performance. K-means clustering is used to group similar instances, and annotators are asked to label a chosen number of points close to the centroids. It assumes one annotator will correctly annotate instances belonging to a cluster, whose centroid has a high-confidence annotation for that annotator. Du and Ling [Du and Ling, 2010] use a single annotator, so the model's judgments are similar to theirs. Consequently, the classification model's decision boundaries are more ambiguous for both the model and the annotator. Wu and co-authors [Wu et al., 2013] use logistic regression to evaluate how well an annotator performs based on the true class labels of instances. It employs the expectation-maximization algorithm that first estimates the true class labels, and then calculates the likelihood of a correct annotation for each class-annotator pair.

Yan and associates [Yan et al., 2011, 2012] use an expectation-maximization algorithm and ground truths to evaluate and estimate the performance of annotators. Zhao and colleagues' [Zhao et al., 2014] model assesses how well annotators perform through two latent variables: the difficulty of the queries and the overall skill of the annotator. If an annotator has high skill or if the query is easy, they are more likely to give the correct answer. These two variables are estimated using the expectation-maximization algorithm. Yang and co-authors [Yang et al., 2018] use a model that learns an embedding for each annotator representing their expertise regarding latent topics, as well as an embedding for each instance related to those topics. Both embeddings are learned through the expectation-maximization algorithm. They are combined to estimate the performance of an annotator. Fang and colleagues [Fang et al., 2012] assess performance as the uncertainty of an annotator into high-level concepts, such as politics, sports, and culture in the case of document classification. An instance may belong to multiple concepts. The model employs the expectation-maximization algorithm to assess the annotator's uncertainty based on instances that are used as ground truths.

Huang and co-authors [Huang et al., 2017] test annotators on ground truths and assume that their performance is the same in similar instances. Thus, to estimate their performance on a new instance, the model references their performance on a similar ground truth. Fang and associates' [Fang et al., 2013, 2014] model assumes that high-level representation of an instance's features and their true class label influence the annotator's performance. This dependency is modeled by incorporating a latent variable that represents the expertise of each annotator, which is then computed as a weighted linear combination of the instance's high-level features. Fang and Zhu [Fang and Zhu, 2014] have annotators labeling instances as "uncertain" when they are unsure. This information is used to assess the annotator's performance by training a classifier that predicts the probability of an instance not being uncertain for that annotator.

Zong, Kading, and respective colleagues [Zhong et al., 2015; Kading et al., 2015] train an annotator model (a support vector machine and Gaussian processes respectively) that predicts if an annotator has sufficient knowledge to annotate an instance or not. This is done by allowing annotators to label instances as "uncertain", which is considered the negative class, unlike providing a label that is the positive class. These instances thus create a binary classification problem. In Chakraborty's work [Chakraborty, 2020], the model requires a labeled pool of data to use as verified ground truths. This allows us to identify the errors made by each annotator. A separate logistic regression model is then trained for each annotator to predict their accuracy as the probability of them correctly labeling new instances. The model presented by Herde and co-authors [Herde et al., 2021b] uses a beta distribution to evaluate an annotator's performance on a specific instance. This evaluation is based on the number of estimated false and true annotations in similar instances, determined by comparing the annotator's work with the predictions made by a classifier trained on other annotators' labels.

## 3.1.2 Emotions and Concentration

As mentioned in Chapter 2, emotions and concentration levels influence a worker's performance. To avoid misclassifications, one must understand how these personal internal factors affect accuracy.

Regarding positivity, Kruger and Dunning [Kruger and Dunning, 1999] state that individuals with positive self-views, and a positive mood, exhibit increased motivation and effort, leading to improved performance. Their performance increases more when offered rewards compared to those in a negative mood. Furthermore, punishment/noxious stimulus leads to avoidance, and decreases performance in similar domains henceforward, enhancing the importance of not punishing workers when they make errors. The paper indicates that a negative mood may reduce motivation, and that when people are in a positive mood (as opposed to negative or neutral), they are more likely to help others. In contexts of collaborative AL, this is an indication that a positive mood might improve results.

While the findings in Kruger and Dunning's paper are relevant, the study analyzed the relation of happiness and performances with cross-sectional studies (meaning, they analyze multiple individuals at one point in time), similarly to many older studies. This approach may have inaccuracies due to unknown individual differences. To assess this problem, more recent works assess variations within individual workers over time. These studies show that longer-lasting moods (not caused by a single stimulus, but rather by consequences of accumulation of experiences) predict future performance, where positive moods improve performance [Koys, 2001; Miner and Glomb, 2010; Rothbard and Wilk, 2011; Staw and Barsade, 1993; Staw et al., 1994]. A drawback of these approaches is that they do not disprove that other external factors could be the reason for this positive correlation. Oswald and his associates, as well as Erez and Isen [Oswald et al., 2015; Erez and Isen, 2002] focus on this issue and still conclude that happiness does improve performance directly.

Several additional studies have also concluded that positive moods resulted in better performance on a given task [Dunker, 1945; Mednick et al., 1964; Kahn and Isen, 1993; Tenney, Poole and Diener, 2016]. Based on this research, one can argue that happier annotators are more accurate. Tenney, Poole, and Diener [Tenney, Poole and Diener, 2016] also explore the correlation between happiness and well-being with other areas (such as health and self-regulation), which seem to decrease fatigue feelings [Behrens, 2022] and increase creativity (which is useful in some tasks).

Additionally, Bellet and colleagues [Bellet et al., 2019] also describe multiple papers that state how happiness correlates with traits useful to work. For instance, individuals induced into positive mood states tend to have more flexible thinking [Isen and Daubman, 1984], be more creative [Isen, Daubman and Nowicki, 1987; Tenney, Poole and Diener, 2016], be more open to information [Estrada, Isen and Young, 1997], be more efficient [Isen and Means, 1983] and have less distractive thoughts [Killingsworth and Gilbert, 2010].

Miner and Blomb [Miner and Blomb, 2010] investigated the relationships between mood and various individual work-related behaviors, including task performance, organizational citizenship behavior, and work withdrawal. They examined how these behaviors fluctuate over the course of the workday. They concluded that positive moods improve performance and that people who can recognize better their moods have a higher boost of performance when in positive moods, as well as a lesser need to take breaks.

To have quantitative values of how much happiness improved performance in a work setting, Bellet and co-authors [Bellet et al., 2019] asked participants how overall happy they felt the past week using a "Face scale", and converted this answer into a 0-to-10 scale. The results show an increase of 12% in productivity by every unit on the former scale, 13.33% increase translating to a 1-to-10 scale. This was a result of happier workers organizing their time better, working faster, and being more efficient at converting calls into sales (without working for longer than usual), compared to less happy employees.

Oswald and colleagues [Oswald et al., 2015] also concluded that an increase of almost one unit on a scale of 1-to-7 leads to 12% more productivity. This translated to a 8% increase when translating the scale to a 1-to-10. The task used to measure this was presented by Niederle and Vesterlund [Niederle and Vesterlund, 2007]. It consists of adding five two-digit numbers for a limited amount of time, and it is a good measure of productivity in jobs where both intellectual ability and effort are rewarded [Oswald et al., 2015]. Their subjects were over a thousand college students from a university with top admission standards. In this case, the authors induced happiness by making participants watch comedy movies or eat sugary snacks. Later, they compared the individuals who got the "happiness treatment" with the control group that did not.

On the other hand, Totev [Totev, 2016], who also used the same test as Niederle and Vesterlund [Niederle and Vesterlund, 2007], conducted the test on Bulgarian kindergarten teachers, on a very similar setup as Oswald et al [Oswald et al., 2015]. Totev concluded that individuals who underwent the "happiness treatment" were 12% more productive by an increase of 1.08 units on a happiness scale of 1-to-7. Translating these values using a scale of 1 to 10, the treatment means an increase of 7.4% for each unit increase in the happiness scale.

Bellet and associates [Bellet et al., 2019] argue that the math-based tasks used in most of these studies are not proven to generalize to real-world work tasks, as these involve other factors such as decisions on how to focus time and energy in multiple tasks. The authors say that in these cases, the correlation between happiness and performance might be smaller. However, as in annotation tasks the workers just annotate sequentially instances that need to be labeled, we can assume these results generalize well for our task.

Oswald and colleagues, Killingsworth and Gilbert [Oswald et al., 2015; Killingsworth and Gilbert, 2010] conclude that happiness increases concentration. Calma and associates [Calma et al., 2016] state (in the context of AL) that human annotators' performance can be influenced by expertise, fatigue, and distractions, and Moran [Moran, 2012] shows that concentration of mental activity is essential for success in any field of skilled performance. Additionally, Caldwell and co-authors [Caldwell et al., 2019] state that repetitive tasks can intensify feelings of fatigue, evidencing the need to not overwork annotators with too many annotations of the same type. The paper studies motor and cognitive fatigue in detail and explains how fatigue reduces decision-making ability, productivity/performance, attention, vigilance, and ability to handle stress on the job. On the other hand, it increases reaction time in speed and thought, forgetfulness, and errors in judgment.

Regarding negative moods, Shackman and colleagues [Shackman et al., 2006] indicate that they may disrupt working memory, Ellis and co-authors [Ellis et al., 1995] state that they seem to reduce comprehension of knowledge, and later [Ellis et al., 1997] indicate that they may decrease cognitive functioning by inhibiting recall of information. Additionally, according to Fredrickson and Branigan [Fredrickson and Branigan, 2005], when compared to a neutral state, positive affects "broaden" thought-action repertoires, while negative affects "narrow" them. Harmon-Jones and associates [Harmon-Jones et al., 2013] show that sadness broadens the attention scope. This contrasted with previous research, such as Gasper and Gerald's [Gasper and Gerald, 2002], that stated that sadness narrowed cognitive scope. Additionally, Rowe and colleagues' [Rowe et al., 2007], did not find a bigger broadening of the attention scope in a negative mood when compared to a neutral mood state. Harmon-Jones and co-authors [Harmon-Jones et al., 2013] concluded that this difference in results comes from how the sadness was induced in the experiments. When only sadness is induced, it broadens the attention scope, but when sadness is felt combined with other negative emotions, it has a negative effect on attention ability.

### 3.1.3 Inability to show uncertainty

Since annotators can be wrong, allowing them to express their doubts decreases the chances of feeding misclassifications to the model. Wu and associates [Wu et al., 2013] used two different methods to calculate the confidence intervals: the Hoeffding's inequality and the Bootstrap method. Then, the authors chose the one that fits their scenario best through experiments. To define the score function they used the uncertainty derived from the learned classifier model and the confidence interval chosen. The study concluded that including the annotators' confidence information, improved the accuracy of integrated labels. Zhong, Fang, and Zhu [Zhong et al., 2015; Fang and Zhu, 2014] also allow annotators to reference if they are uncertain about how to label a class and use this information to predict how good an annotator is for a specific type of class.

Another way to manage uncertainty is to allow gradual annotations. Calma and colleagues [Calma et al., 2018b] merge per annotation the class label given by an annotator, and a self-assessed confidence regarding that answer, creating a gradual annotation. They concluded that as the number of misclassified instances increases, the advantages of using gradual labels also increase, confirming the idea that presenting uncertainty is especially important when, otherwise, misclassifications would enter the model. Sandrock and co-authors [Sandrock et al., 2019] combine 3 types of confidence: human-inspired confidence, machine-generated confidence, and non-normalized confidence. They combine these confidence scores in two fusion strategies for combining confidences provided by multiple annotators - "Poisson Binomial Based Confidence Fusion" and "Incomplete Beta Function Based Confidence Fusion". These strategies help with the problem of inter-annotator agreement, finding the label selected as true taking into account confidence scores.

Multiple studies show that annotators are usually reliable in their confidence scores [Wallace et al., 2011; Calma et al., 2018a], but the Dunning-Kruger-effect [Kruger and Dunning, 1999] shows that unskilled annotators tend to overestimate their skills, not recognizing their limits [Kruger and Dunning, 1999; Gadiraju et al., 2017]. Additionally, very skilled experts tend to underestimate their ability and test performance relative to their peers [Kruger and Dunning, 1999].

To avoid uncertainty unrelated to query difficulty, the task description must be clear and not ambiguous. The annotators should know what is expected from them and not have doubts about functionality or course of action [Munro, 2021]. Hossfeld, Tokarchuk, Georgescu, and respective co-authors [Hossfeld et al., 2014; Tokarchuk et al., 2012; Georgescu et al., 2012] state that clarity is positively correlated with performance on the task. Therefore, if an instance is not clear or the annotator does not correctly understand what is expected from them, they will be more confused and uncertain, leading to more misclassifications.

# 3.2 Query-annotator pairs

Zhong and associates [Zhong et al., 2015] show how considering an annotator's performance to choose which worker annotates an instance benefits the learner queries, as that way, we are sure to know if there is at least one annotator skilled enough to answer it correctly. Herde and co-authors [Herde et al., 2021a] describe two different strategies that have been used to define these query-annotator pairs:

- Sequential selection - A ranking of annotators is done regarding a query. In this approach, relative performance is more important than an exact quantitative quality of each annotator. One might predefined a number of annotators with the highest estimated performances per query [Long et al., 2013];
- Join selection - It is based on a selection algorithm that simultaneously evaluates both the queries and the annotators. The utility of each query and the performance metrics of the annotators must be effectively combined. One common method for achieving this is by calculating the product of the two metrics as done by Huang [Huang et al., 2017].

**Sequential Selection**

Considering sequential selection, Fang, Rodrigues, Wu, Zhong, and respective colleagues [Fang et al., 2013, 2014; Rodrigues et al., 2014; Wu et al., 2013; Zhong et al., 2015] select the query with the highest estimated utility and the annotator with the highest estimated performance when labeling that specific query. Fand and associates [Fang et al., 2012] use this methodology too, but the annotation with the lowest estimated performance is also chosen, so that the expert in the query can teach them, as a form of collaboration between annotators. Similarly, Calma, Long, and co-authors [Calma et al., 2018a; Long et al., 2016; Long and Hua, 2015] adopt this approach. However, instead of selecting a single annotator, they select multiple annotators for the labeling process. Yang and colleagues [Yang et al., 2018] on the other hand select multiple queries with the highest utility measure, and then each to the respective annotator with the highest estimated performance.

Wallace and associates [Wallace et al., 2011] explore a multiple expert AL (MEAL) scenario. The query with the highest estimated utility is selected, and a categorical distribution is used to choose the appropriate annotator. It is an approach especially useful when using expert annotators, who are being paid even when not annotating, as the distribution used is parameterized to achieve specific goals, such as ensuring an equitable distribution of workload.

Zheng and co-authors [Zheng et al., 2010] select the query with the highest estimated utility. chooses an adaptive number of annotators with the highest estimated performances to the chosen query, and subsequently selects a fixed subset of annotators with low annotation cost and high performances.

**Join selection**

Considering join selection, Donmez, Moon, Huang, and respective colleagues [Donmez et al., 2008; Donmez and Carbonell, 2010; Moon and Carbonell, 2014; Huang et al., 2017] choose the query-annotator pair that has the highest product of estimated query utility and annotator performance. Yan and associates [Yan et al., 2011] try to find the perfect trade-off between the most useful query and a high-performance annotator. Nguyen, Herde, and respective co-authors [Nguyen et al., 2015; Herde et al., 2021b] jointly select a query and an annotator by integrating the estimated performance of an annotator (or group of annotators) into the utility measure of the query, quantifying the performance gain of the classification model. In order to select a batch of query-annotator pairs Chakraborty [Chakraborty, 2020] aims to balance the trade-off among useful queries, accurate annotators, and minimal redundancy between queries by solving a linear programming problem.

## 3.3 Active Learning and Recommendation Systems

While recommender systems have never been used to improve AL itself in our knowledge, AL has been used to improve RS.

The first time this combination occurred, the idea was to improve sign-up processes Desrosiers and Karypis [Desrosiers and Karypis, 2011]. At this point, the system gives one or multiple movies for the user to rate. The decision of which movies to choose for this uses AL, aiming to get the ratings of movies that improve the RS the most [Elahi et al., 2016]. To illustrate using Figure 3.1, imagine an app with four types of movies: "a", " b", "c" and "d". Presenting the only movie in "a" to get an initial rating is not very useful for making predictions on the best future recommendations for the user, as it is a peculiar movie. On the contrary, if the system chooses a movie in the "d" category, as there are many similar movies, the RS can predict the user's rating on multiple movies, being the ideal choice.

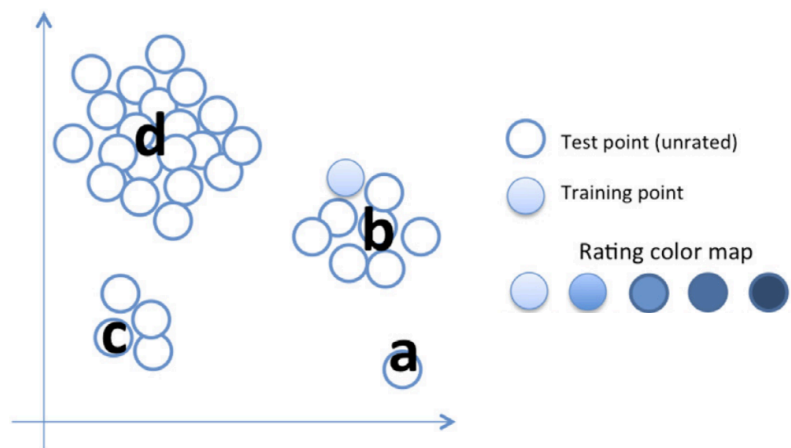

Figure 3.1 - An example of AL's use in RSs. Adapted from [Elahi et al., 2016].

Yang and colleagues [Yang et al., 2020] explore the use of a function that balances the use of RS or AL to make predictions about the user, introducing a parameter ($\alpha$) that allows for interpolation between pure recommendation and AL. For recommendation, the framework selects items that maximize the probability of being relevant, while for AL, it focuses on items that reduce uncertainty about their relevance. The paper did not find the AL addition to be very advantageous, getting higher performance in both recommendation accuracy and predictive accuracy when using purely recommendation.

The cold-start problem is a common drawback in some types of RS, that happens when the system does not have enough ratings to make good recommendations [Jannach et al., 2010, Pozo, 2018]. AL has also been used by multiple authors as a way to overcome this limitation of RSs [Bu et al., 2018; Zhu et al., 2020; Geurts, 2020; Pozo, 2018], and the results were positive. This solution is possible by using AL to obtain data that better represents the users' preferences, which improves the performance of the RS and addresses the cold-start problem [Elahi et al., 2016].

Chen and colleagues [Chen et al., 2021] formalized an RS into a Positive-Unlabeled learning [Denis, 1998] task (where only positive or unlabeled data is used, and which has been proved to perform better than models trained using conventional binary classification learning algorithms after doing negative instance sampling [Chen et al., 2021]). AL is used in this paper as a way to solve the problem of insufficient labeled data. This combined approach presented better results by reducing Positive-Unlabeled learning's query cost.

Ko and co-authors [Ko et al., 2022] present a survey of RS, where various application fields are mentioned, such as social network services SNS, streaming services of music and video, tourism field (to recommend tourist destinations), health, and others. The paper describes multiple recent applications of RS from the three presented types in the previous chapter. Many of these include the use of neural networks, clustering, and other RS techniques, in order to improve recommendations. In this paper, most RS used in health fields is CFB RS, as it selects health content based on personal data, ensuring personalized health information tailored to their specific needs [Abbas et al., 2020; Sanchez et al. 2017; Wiesner and Pfeifer, 2014].

## 3.4 Implications and Insights to our work

The previous sections provide an overview of the state-of-the-art literature, which contributes with implications and insights for the present work.

This work focuses on multiple annotators instead of just one, as the main goal is to create a way to distribute instances to label among the multiple annotators available, taking into account their individual factors (such as average accuracy, fatigue level, and mood). This work follows the query-dependent annotator performance type, and, in a comparable manner to Huang and colleagues [Huang et al., 2017], it assumes that annotators have a similar average accuracy in similar instances.

According to many authors referenced in the previous section [Bellet et al., 2019; Caldwell et al., 2019; Calma et al., 2016; Dunker, 1945; Erez and Isen, 2002; Fredrickson and Branigan, 2005; Kahn and Isen, 1993; Killingsworth and Gilbert, 2010; Mednick et al., 1964; Miner and Blomb, 2010; Moran, 2012; Oswald et al., 2015; Tenney, Poole and Diener, 2016] less fatigue and/or positive mood states are positively correlated with performance. Therefore, similarly to how Wallace and peers [Wallace et al., 2011] assumed that higher pay indicates better performance, this work will assume that better mood and less fatigue indicate better performance.

To provide higher confidence in the assumptions that will be used, it is important to draw upon literature specifically related to situations similar to classification and annotation tasks. Calma and colleagues [Calma et al., 2016] present a paper relevant in an AL context, while Fredrickson and Branigan's research [Fredrickson and Branigan, 2005] includes an experiment involving a decision-making task between two options, which can be viewed as a classification task. These papers conclude that fatigue negatively impacts annotators' performance in the context of AL and that positive emotions (more intense - e.g. amusement - and less intense - e.g. contentment) broaden the scope of attention and thought-action repertoires. As these statements align with the findings of other research discussed in Section 3.1.1, and align with the task of this work, these conclusions (also shared with other studies), will serve as the foundation for this work. Consequently, this work considers that positive emotions lead to better performance than negative or neutral emotions and that more fatigued workers will perform worse compared to days when they are not fatigued.

Bellet, Oswald, and their respective co-authors [Bellet et al., 2019; Oswald et al., 2015] concluded that there was a 12% increase in productivity when happiness increased. Neither of the two studies was conducted in a classification/annotation context (which would be similar to that of this work), therefore, it can not be assumed that such a significant increase in accuracy will happen. Nevertheless, productivity might be measured in different ways, such as efficiency and accuracy [Sink et al., 1984], which allows the claim that part of the 12% includes speed when deciding labels, and not in the accuracy itself. For that reason, being conservative, it will be assumed that each unit increase in a 1-to-10 scale of mood (referencing Bellet and colleagues [Bellet et al., 2019]) generates an increase of 6% in the annotator's accuracy. The average accuracy of the annotator will be connected to their average mood evaluation on the 1-to-10 scale.

CFB RS is effective in health fields [Abbas et al., 2020; Sanchez et al., 2017; Wiesner and Pfeifer, 2014] due to how it selects individual data for health content selection, ensuring tailored information. Similarly, this work will use a Knowledge-Based RS to select annotators for each instance. This approach ensures that the best annotator for each specific query is selected when considering multiple factors. These factors include the annotator's past accuracy, and their current levels of mood and fatigue.

# Chapter 4
# Methodology

This chapter offers a detailed description of the experiments carried out in this study, addressing key elements like dataset preparation, parameter selection, and additional relevant factors.

## 4.1 Approach

To answer the research questions of this work, several critical steps are undertaken. The AL task is a query-dependent annotator performance type (assuming the annotator's performance depends on the characteristics and skill of the annotator, the complexity of the annotation, and the nature of the query in question). The characteristics of the annotator that is taken into account are the following:

- Mood - This assessment is done with a 1-to-10 scale that can change every section of work time. On this scale, 1 indicates a very bad mood, while 10 indicates a very good mood. The average value for each annotator is assumed to be the unit value that makes the annotator perform at their average accuracy. This average mood will be between 3 and 7 to avoid extreme rare cases, being realistic;
- Fatigue - This analysis is done by looking at how many queries have been labeled by the annotator in two consecutive periods of the same day. A threshold of 50 is used as the starting point to consider the annotator as fatigued (level one). For each 20 more annotations, the fatigue level increases by 1 unit. An increase in level is indicative of an increase in fatigue;
- Past average accuracy - This value is artificially created when the simulated annotators are created (see below). There is an average accuracy and an accuracy for each possible label to represent historical values;
- Chronotype - This information is used to simulate variances in the mood throughout the day that follow the literature aforepresented. It is simulated using distributions on the population, regarding sex and age.

To define each query instance, the following features are used:

- Query difficulty - This assessment is based on the model's uncertainty when trying to label an instance. Therefore, the higher the entropy of the AL model's prediction, the greater the perceived difficulty of the query;
- Query type - This assessment is based on the label(s) that the model is more confident in. For instance, consider an AL situation that has three possible labels, and an instance where the model's prediction probabilities for label A, B and C are 45%, 35% and 20% respectively. In this case, the query type would be A and B, as those are the labels labels whose predicted probability exceeds the reciprocal of the number of labels (in this case, $\frac{1}{3} \times 100$).

Having the information about the query type, the RS can check for the past accuracy of each annotator on those labels, and on average for all their annotations, as well as current mood and fatigue level to help make the best recommendation.

The RS used to define the query-annotator pair is similar to a Knowledge-based RS. In this context, making the comparison between traditional RS usage, the instances function as users, while the annotators represent the items being recommended. We use weights considering knowledge about the items (annotators) and the user request (in this case, the instance). Based on the instance to be labeled, each annotator is considered, achieving a score based on weight sums using knowledge from both parts. In the end of the process, the annotator with the biggest score (similarity) is chosen (recommended).

This work considers a working day to be divided into three periods (simulating the time from 8 AM to 11 AM, from 11 AM to 3 PM and, finally, from 3PM to 6PM). These periods allow us to show how mood and performance varies through the day, as well as consider how tired a person is. For that, we consider how much they worked in the current, and previous (if applicable) period of that day. For instance, if the annotation occurs during the third period we consider the number of annotations performed by that annotator in the current period, and in the second period.

We use numbers of annotations to simulate how each period and day ends. Each period lasts for 204 annotations, meaning each day lasts 816 annotations. As we could not find empirical data on how annotations are usually done in a setting similar to ours, the number 204 is heuristic. This number is high enough to see how differences in mood levels through the day affect the performance of the annotators, and low enough that, given the implementation of fatigue in our model, avoids prolonged effects that would be inconsistent with realistic human fatigue dynamics. In the future, it is important to test different values, and explore how to simulate the passing of a day in a more realistic way (for instance, considering how annotations may happen in parallel, including the factor of time efficiency).

The study is divided into four tests: one test to compare the use of past performance and mood, another for past performance, fatigue and mood, the traditional AL approach (which only considers past accuracy), and one test of optimization. This fourth test is used to compare how good our proposed approach is compared to the best option for the experimental setup created. It will mainly serve as a comparison point.

A weighted aggregation is used by the RS to define the query-annotator pairs, using the features considered in that test. Annotators available are ranked based on their best compatibility with the query, based on past and current performance (which considers mood or/and fatigue values. The annotator with the highest score is chosen to annotate the query for the active learner. This way, when doing the test representative of our proposed approach, the system takes into account not only the annotator's skill, but also their well-being (as it influences their productivity) and distribution of work (which is related to fatigue levels) to find the best query-annotator pair.

The original idea was to create a unique and more efficient way of selecting annotators to label the instances that give more information to the model. The AL model using this RS as aid aims to achieve higher accuracy more quickly by selecting annotators based on their reliability for specific instances, and performance capability at the moment, rather than relying solely on past accuracy.

## 4.2 Data, annotator simulation and active learner

To simulate the AL setting we had to decide on which datasets to use, and on many parameters needed. These decisions were based on science, on realistic observable values and on what could be used to explore the results with more depth.

### 4.2.1 Data

Concerning the datasets employed in this research, a diverse range was selected to assess the applicability of the proposed methodology across different contexts. The datasets selected for this study are as follows:

- Breast cancer detection [Wolberg and Mangasarian, 1993]: A small dataset based on numerical features commonly used for binary classification tasks.
- Wine quality prediction [Cortez and Cerdeira, 2009]: A medium-sized tabular dataset where the possible wine quality scores range from 0 to 10;
- MNIST digit classification [Li, 2012]: A large image dataset with 70,000 grayscale images of handwritten digits with 10 possible labels (digits 0–9);
- Fashion MNIST classification [Han et al., 2017]: A large image dataset with 70,000 grayscale images of fashion item categories. Similarly to MNIST dataset, it has 10 possible labels;
- Titanic survival prediction: A small tabular small dataset with mixed data types (categorical and numeric) with binary classification.

### 4.2.2 Annotator-related parameters based on research

Focusing on parameters, many were modeled to realistically represent the annotators. Given the resource constraints and the complexities involved with using real annotators, the decision was made to develop simulated annotators instead. The simulated annotators were designed to mirror real-world data distributions based on established research documenting distributions of human characteristics presented in the previous chapters. For this reason, the following decisions were made:

- Age and gender: To simulate many real life distributions of age in workers, these values are changed in each batch of annotators. This variation is done with probability of bellowing in four age groups:
    - 25 to 37-years-olds;
    - 38 to 45-years-olds;
    - 46 to 55-years-olds;
    - 55 to 65-years-olds.

  The division of age groups was done based on Fischer, Sládek and co-authors' work [Fischer et al., 2017; Sládek et al., 2020] regarding chronotypes. These are the age groups where more changes happen. It starts with a fast decrease in chronotype with age, then stabilization, followed by an inversion between males and females, finishing with a decrease in chronotype again for both genders.

- Chronotypes: To simulate the hours where each annotator is most productive and the probability of having each chronotype. The approach presented by Breus [Breus, 2016] referenced in Chapter 2 is used for the relationship of hours of day and productivity for each chronotype. Furthermore, differences in this variable due to age are accounted for by following Fischer, Sládek and colleagues [Fischer et al., 2017; Sládek et al., 2020] considerations. For this reason, we consider different probabilities for being each of the four chronotypes presented by Breus [Breus, 2016] depending on the age group.

- Being a "Bear" is close to 50% in all age groups;
- Being a "Dolphin" is very rare for people over 45 years old, but 10% likely between 38 and 45 and 20% likely between 25 and 37-year-olds;
- Being a "Wolf" is 25% likely between 25 and 45-year-olds but after that only 15% likely;
- Being a "Lion" is much less likely with younger ages, making it 10-15% likely between 25 and 45-year-olds, while being around 35% likely between the age group of 46 and 65-year-olds.

The probability's specific values are intended to follow the aforementioned work closely. To simulate how chronotype affects productivity at different times of the day, we connect it with mood values, as awakeful levels correlate with mood values [Kline et al., 2010; Koorengevel et al., 2003; Murray et al., 2002]. The details will be further explained in the section about fatigue.

- Mood: To simulate real mood values we start by assigning an average mood between 3 and 7 on a scale of 1-to-10, avoiding the extreme values. After that, we account for daily fluctuations according to someone's chronotype. For instance, "Lions" have a higher probability of being happy in the morning than "Dolphins", while the opposite is also true for the last few hours of the work day [Breus, 2016; Fischer et al., 2017; Sládek et al., 2020].

### 4.2.3 Annotators' performance parameters

Regarding average performance, we use a random value from a normal distribution with a mean of 75 and a standard deviation of 7, in order to simulate different sets of annotators that may be found in real life. The values of the mean and standard deviation do not follow any research, as none was found about these values on annotation tasks.

Furthermore, the performance of each annotator in different periods of the day is assumed to be affected by mood and fatigue. To simulate this, we follow the literature, seeing how much productivity is affected by mood. We consider units of difference between each annotator's average mood, and the mood value in each period of the day in the 1-to-10 scale. As aforementioned, studies [Oswald et al., 2015; Totev, 2016] concluded that a unit increase on a 1-to-10 scale of mood led to around 8% increase in productivity. As this percentage already takes into account both speed and accuracy, we consider a more conservative value. This adjustment is necessary because we account for the effects of fatigue by the variable regarding fatigue (which takes into account speed too), not together with mood. As fatigue makes humans slower and more prone to errors, it would not be correct to use the full value of 8% indicated in the literature. For this reason, we assume that each unit of mood difference from the average results in a corresponding 6% change in the specific period: an increase for upward differences and a decrease for downward differences.

For the initial annotator's performance for each label, we use a random value from a normal distribution whose mean is each annotator's overall performance and a standard deviation of 6 (to be smaller than the original one for each annotator average value). The aim is to simulate natural differences in performance for different labels.

During the annotation process, performance values for each label, and the mean is updated after each annotation day has ended (3 periods). The starting values are assumed to come from 100 annotations of each label in the past.

### 4.2.4 Fatigue consideration in this work

Concerning the affects of fatigue, there are two factors to consider. First, each chronotype has a period of four hours in the day where each is more productive according to Breus's work [Breus, 2016], concerning circadian rhythm. Secondly, there is the fatigue caused by actual work done.

To account for the influence of individual chronotypes on productivity, we considered their connection with mood, as aforementioned. We follow the conclusion: higher wakeful levels are correlated with better moods, which are correlated with better performance [Kline et al., 2010; Koorengevel et al., 2003; Murray et al., 2002, Oswald et al., 2015, Totev, 2016]. As indicated above, these experiences consider simulated working days divided in three periods. These periods don't follow exactly the time windows provided by the boost in productivity assessed by Breus [Breus, 2016], but it remains feasible to navigate this discrepancy.

We make sure Lions are more likely to be happy in the first two periods, compared to all the other types (as their most productive hours are from 9 AM to 2 AM), and less likely to be as happy as the other types in the last period. As Dolphins have a productive window from 3PM to 7PM we make sure they get increasingly likely to be happier through the day, especially at the last period. Finally, for both Bears and Wolves we just make them more likely to be happier in the last two periods (as their most productive hours are from 10 AM to 2 PM and 1 PM to 5 PM respectively).

The average mood for the first period is a random integer between 1 and the annotator's average mood (in case of Lions, we consider the average mood plus one to make them more likely to be happier in the mornings). We use this value from the first period as the basis of how the mood fluctuates through the day. An example of how the value of mood is generated for a "Dolphin" annotator through for a day is represented in Figure 4.1. Through this variation of mood values (ensuring that they follow a realistic pattern through the day), we can simulate how their chronotype affects their productivity across different periods [Murray et al., 2002].

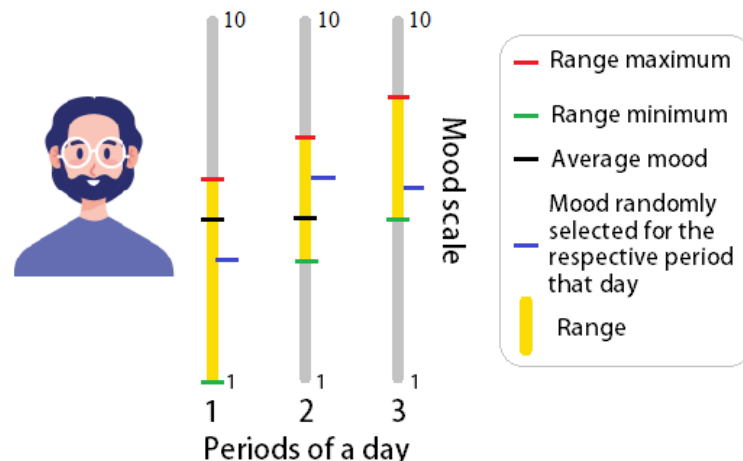

Figure 4.1 - An example of the simulation of mood variation
for an annotator with "Dolphin" chronotype.

Beyond chronotype, the second factor we considered regarding fatigue was how it results from the work itself. Qualitative values on how fatigue affects the performance in specific tasks were not found in the sources we reviewed. In the absence of such data, we were required to assume a value in order to carry out the analysis. We considered two plausible values based on heuristic reasoning (2% and 4%) and compared the results. The reasoning is also based on the disconsideration of the complete productivity advantage of an unit increase in mood scale, as aforementioned. The use of two values is not to examine the sensitivity of the results to this parameter, but rather as a necessary step to enable the analysis, demanding the examination of different values.

This percentage is deducted from the annotator's actual labeling performance in each threshold. The initial threshold is set at 50 annotations, followed by recurring thresholds at every additional 20 annotations. These numbers were chosen based on the amount of annotations done per period. The annotations considered to determine fatigue levels are the last, or two last, consecutive periods of the same day. We consider 50 annotations to be a reasonable point at which annotators may begin to experience fatigue during 6 hours of work, as this represents about one-eight of the total annotations completed over two periods. It should be noted that, while in a real life setting a smaller number of annotators would require more time to label the same number of instances as a larger group, this study does not consider time efficiency as a variable, as it is not the focus of this study. This consideration would justify changes on how time is considered, and on how many labels are used for fatigue levels.

Figure 4.2 illustrates how the framework of fatigue levels works for one annotator during one day, when their performance worsens by 2% at each threshold. The initial predefined performance values are created as aforementioned, basing themselves on the given overall performance of the annotator, but already affected by mood changes each period and day.

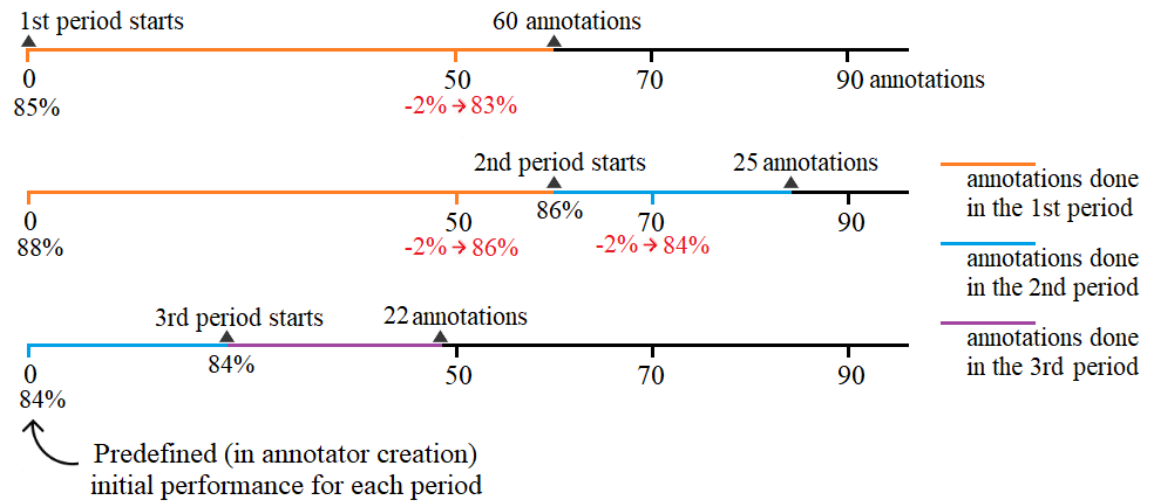

Figure 4.2 - An example of determining fatigue levels
of an annotator during a working day

## 4.2.5 Active learner

Referring to the active learner, it was initialized with a random forest classifier as the underlying estimator. Random forests represent a category of techniques that involve creating an ensemble of decision trees that are developed using a modified version of the decision tree learning process. In this approach, each tree in the forest is generated from a different subset of the training data and utilizes random selections of features, which helps to reduce overfitting and improve overall model accuracy [Louppe, 2015].

The stopping criteria for the active learner was a number of annotations 1224 annotations performed, or an accuracy of 99% (traditional approach). The number 1224 was chosen because it covers two full days, or six periods. This allows for the influence of different moods and fatigue levels on the annotators' performance to be captured. At this point, the model's learning curve has also stabilized, we have a number of annotations enough to compare test results, considering the dataset size.

## 4.3 Recommendation System

The Knowledge-Based RS uses the aforementioned attributes to create a rank of annotators. Algorithm 1 represents how it is implemented. It takes into account the labels that confuse the model the most, and uses only those to check the average performance of each annotator in it, as well as using the overall average performance in all labels. The weights considered for the overall accuracy, and the accuracy on these labels change depending on the characteristics of the instance. Furthermore, the way the performance is calculated may only include the value of the past accuracies in the case of the Test 1, consider additionally the mood values in case of Test 2, and consider accuracy, mood and fatigue levels in case of Test 3. This is the coding difference between the three main tests (excluding the optimization-based approach).

---

**Algorithm 1** Pseudo-code for the Knowledge-Based RS

---

1: **Given:**
2:     $u$ - uncertainty of the queried instance
3:     $L$ - list of the active learner's probability for each label
4:     $Anot$ - Dataset with information about available annotators(including average accuracies and mood values)
5:     $N$ - dictionary with the number of annotations done for each annotator in the considered time frame for fatigue.
6:     $sd$ - mean value of model uncertainty for +2 standard deviations
7: collect $t$ labels whose probability is greater than 1 divided by the number of labels
8: create empty list $scores$
9: **for** each annotator in $Anot$ **do**
10:     $score = 0$
11:     get $ac$ RS's predicted accuracy in general as aforementioned using $Anot$ and $N$ (considers only overall performance in test 1; performance and mood levels on test 2; performance, mood and fatigue levels on test 3)
12:     **if** $u > sd$ **then**
13:         set $w = 0.8$ weight used for the overall predicted accuracy and $w_l = 0.2$ weight for the predicted accuracy on $t$ labels
14:     **else**
15:         **if** size of $t$ > half of the size of $L$) - 1 **then**
16:             set $w = 0.3$ weight used for the overall predicted accuracy and $w_l = 0.7$ weight for the predicted accuracy on $t$ labels
17:         **else**
18:             set $w = 0.5$ weight used for the overall predicted accuracy and $w_l = 0.5$ weight for the predicted accuracy on $t$ labels
19:         **end if**
20:     **end if**

21:     **for** each label $i$ in $t$ **do**
22:         get $ac_i$ RS's predicted accuracy for each label (considers only overall performance in test 1; performance and mood levels on test 2; performance, mood and fatigue levels on test 3)
23:         set $score_w = w \times ac + w_l \times ac_i$
24:         update $score$ += $score_w$ * $L[i]$
25:     **end for**
26:     Append $score$ to list $scores$
27: **end for**
28: Sort list $scores$ descending
29: Get the top index of $scores$

---

Algorithm 1 - Knowledge-Based RS proposed

## 4.4 Experiment design

All tests done in this study are done on three different batches of 30 annotators. They were created as aforementioned, and they are all always available during the training process. The choice of the number of annotators per batch was heuristic, as annotators' team sizes vary for different projects, and there is no rule of thumb or data to decide on a number. Having 30 annotators allows us to have enough annotators to be able to make conclusions, but it is also fitted for the amount of labeling done with the datasets used. The use of three batches is to compare how much the proposed approach is really helpful, as it helps to solidify the results. This way we have a lot of variability in the annotators and can use the tests in three different situations. In Appendix A it is visible the distributions of the variables aforementioned regarding the three batches.

This study includes four tests:

- Test 1: Use of the RS, which only considers each annotator's past accuracy. This test serves as the comparison point between the proposed approach (Test 3) and traditional AL techniques (that only consider past performance);
- Test 2: Use of the RS, which considers each annotator's past accuracy and mood levels. This test will let us compare the proposed approach with one that is easier to implement, considering only how the emotional state influence one's performance;
- Test 3: Use of the RS, which considers each annotator's past accuracy, mood and fatigue levels. This is the proposed approach of this thesis. We use this test in comparison to Test 1 to answer the Research Question 1 and 2;
- Test 4: Use of an optimization approach, based on how the annotators label the instances. This test allows us to answer the Research Question 4.

Tests 1, 2 and 3 vary in the way they predict the annotator's accuracy at that moment.

- For Test 1, only the past accuracy of the annotator is used. Their mood and fatigue levels are not taken into account in any way.
- For Test 2, the predicted accuracy of the annotator at the time of the query considers past accuracy values and how the annotator's mood affects their performance at the moment.. This means that it considers how mood varies between periods, showing it through different predicted accuracy;
- For Test 3, the predicted accuracy of the annotator also considers fatigue on top of the predicted accuracy for Test 2. For that, we consider how many annotations the annotator did that day in that period and in the period before (if plausible), and decrease the predicted accuracy aforementioned by 2% or 4% at each threshold value of annotations.

Regarding Test 4, the algorithm used can be seen in Algorithm 2. As aforementioned, the optimization process bases itself completely in the code for the simulation of labeling by the annotators. Instead of making the annotator chosen by the RS perform based on the code, the optimization code goes through all the available annotators. It chooses the one whose predicted accuracy for the given query is the highest, based on the labeling process. This approach is the best possible scenario given how the experiments are made, as it bases itself on how the annotators are simulated to label the instances.

---

**Algorithm 2** Optimization process

---

1: **Given:**
2:     $u$ - uncertainty of the queried instance
3:     $L$ - list of the active learner's probability for each label
4:     *Anot* - Dataset with information about available annotators(including average accuracies and mood values)
5:     $N$ - dictionary with the number of annotations done for each annotator in the considered time frame for fatigue.
6: Collect and sort in descending order the $t$ labels whose probability $L$ is greater than 1 divided by the number of labels
7: Create an empty list *scores*
8: **for** Each annotator available *Anot* **do**
9:     From $N$ get the current number of annotations of the annotator in the last or, if available, two last consecutive period $n_{annotations}$
10:     Create list *accuracyLabels*
11:     **for** Each label in $t$ **do**
12:         Append the annotator's performance for that label on *accuracyLabels*
13:     **end for**
14:     Sort *accuracyLabels* descending
15:     **if** Size of *accuracyLabels* $< 3$ **then**
16:         Append 0 on *accuracyLabels*
17:     **end if**
18:     Set *meanAccuracy* as the predicted overall accuracy of the annotator using *Anot* (which includes the influence of mood)
19:     Apply the fatigue levels for *meanAccuracy* and for each element of *accuracyLabels* as aforementioned, with a minimum value of 0
20:     **if** size of $t$ >1 and $accuracyLabels[0] - accuracyLabels[1] < 0.04$ and $accuracyLabels[1] - accuracyLabels[2] >= 0.04$ **then**
21:         set $w_{highest} = 0.3$, $w_{2ndhighest} = 0.3$, $w_{3rdhighest} = 0$, $w_{meanAccuracy} = 0.4$
22:     **end if**
23:     **if** $|t| > 1$ and $accuracyLabels[0] - accuracyLabels[1] < 0.04$ and $accuracyLabels[1] - accuracyLabels[2] < 0.04$ **then**
24:         set $w_{highest} = 0.3$, $w_{2ndhighest} = 0.2$, $w_{3rdhighest} = 0.2$, $w_{meanAccuracy} = 0.3$
25:     **else**
26:         set $w_{highest} = 0.5$, $w_{2ndhighest} = 0$, $w_{3rdhighest} = 0$, $w_{meanAccuracy} = 0.5$
27:     **end if**

28:     Set *score* as a weighted average: $accuracyLabels[0] \cdot w_{highest} +$
        $accuracyLabels[1] \cdot w_{2ndhighest} + accuracyLabels[2] \cdot w_{3rdhighest} + meanAccuracy \cdot$
        $w_{meanAccuracy}$
29:     Append *score* to *scores*
30: **end for**
31: Sort list *scores* descending
32: Get the top index of *scores*

Algorithm 2 - Optimization-based approach

After either Algorithm 1 (in case of Test 1, 2 or 3), or Algorithm 2 (in case of Test 4) are run, the chosen annotator labels the instance. The uncertainty of the model for each queried instance, the accuracy of the AL until that moment, the chosen annotator and the information of if the annotator was right or wrong when labeling the queried instance are saved. This information will allow us to show the learning curve of the algorithm, and also check how many correct annotations are done through the training process in each test. We also note the CPU and execution times.

## 4.4 Chapter summary

This chapter provides a description of the experiments conducted in the study and the outline of the approach, covering aspects such as the datasets used, the parameters used for simulating the annotators in a realistic way, and presenting the active learner and RS.

We explain how mood and fatigue levels are considered in our approach, and present the four tests performed in this work, aiming to answer its Research Questions. The background literature used to support choices in parameters is presented, as well as the RS's and the optimization-based approach algorithms.

Each test performed simulates two days of work for the annotators, which include 3 periods each. Their mood values can vary each period based on their average and their original mood that day, while their fatigue levels increase when reaching thresholds of annotations performed.

# Chapter 5
# Results and Discussion

In this chapter, we present and analyse the results of the experiments, comparing the results obtained from the different methods.

## 5.1 Results

The results of the five datasets used in this research are presented in this section. We present the average results of all the values of how many instances were correctly labeled by the selected annotators, accuracy and F1-score of the model through the training phase, and its uncertainty on the query instances in each approach. Appendix B includes the results of F1-score (which followed accuracy very closely, not needing to be presented in the main text). Learning curves are presented with smoothness for better visualization.

Starting with the tests done using 2% decrease per fatigue level, we will present the values for the wine quality prediction, breast cancer detection, Titanic survival prediction, MNIST digit classification and Fashion MNIST classification in this order. We continue by presenting the average of all tests and batches of annotators, finishing with an overview of the results, and presentation of CPU and execution times.

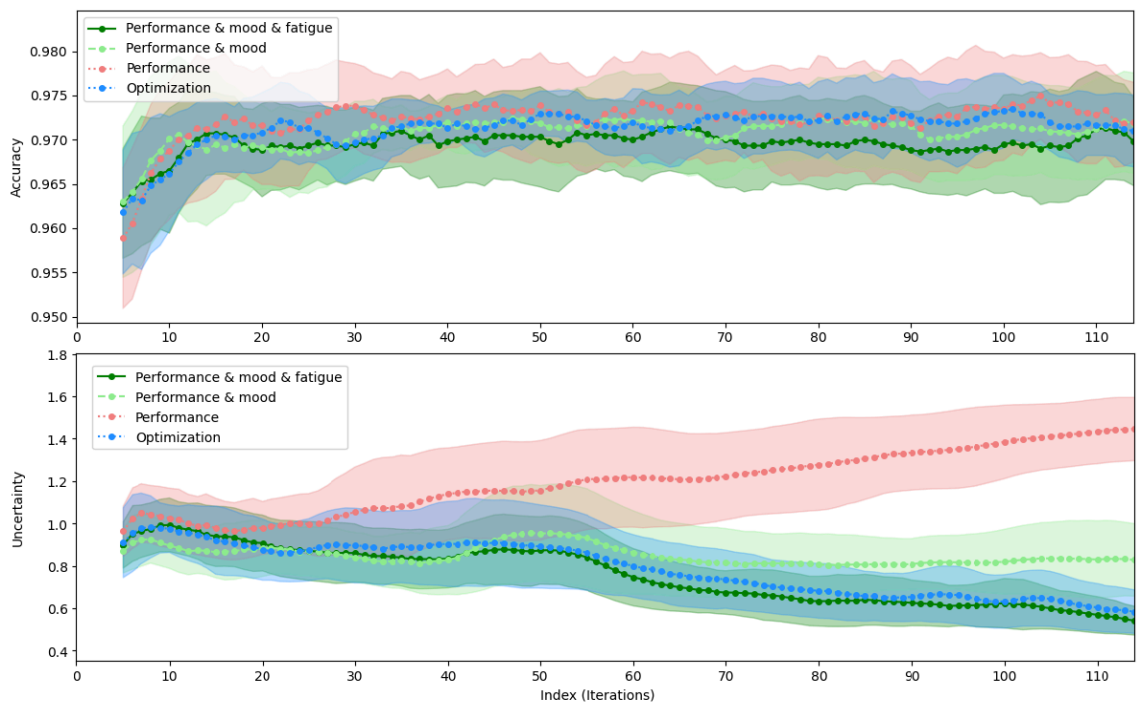

Figure 5.1 - Mean values of model accuracy on full dataset while training, and uncertainty on queried instances for wine quality prediction dataset with fatigue levels worsening performance by 2%

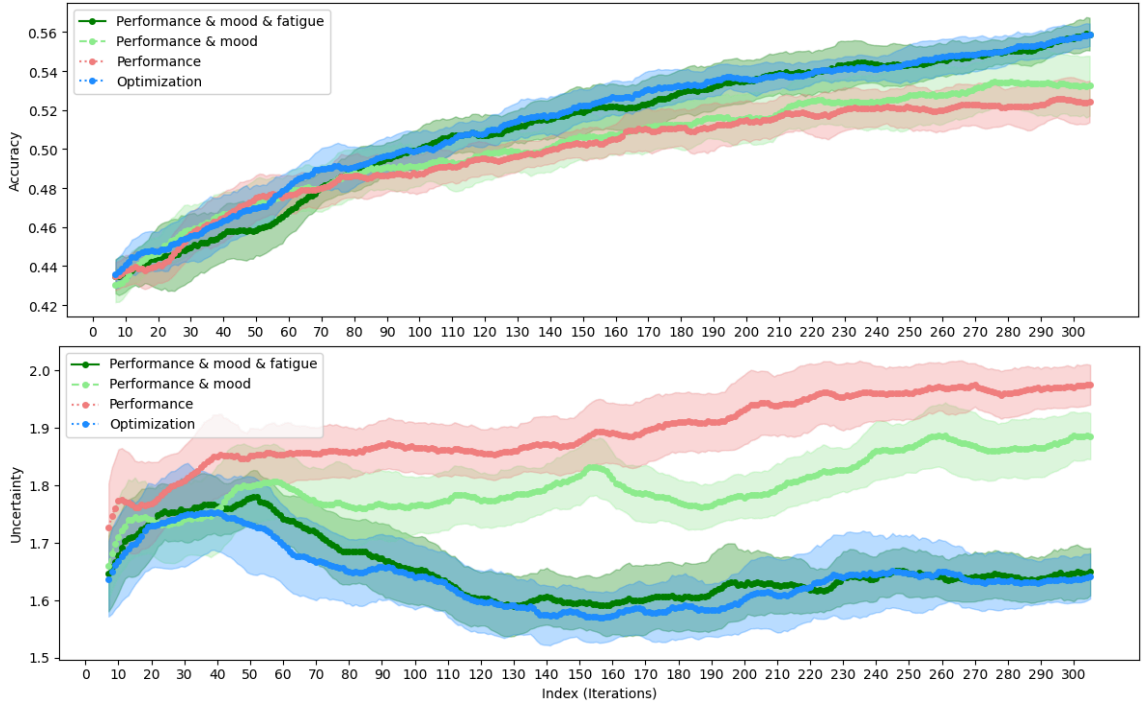

Figure 5.2 - Mean values of model accuracy on full dataset while training, and uncertainty on queried instances for breast cancer detection dataset with fatigue levels worsening performance by 2%

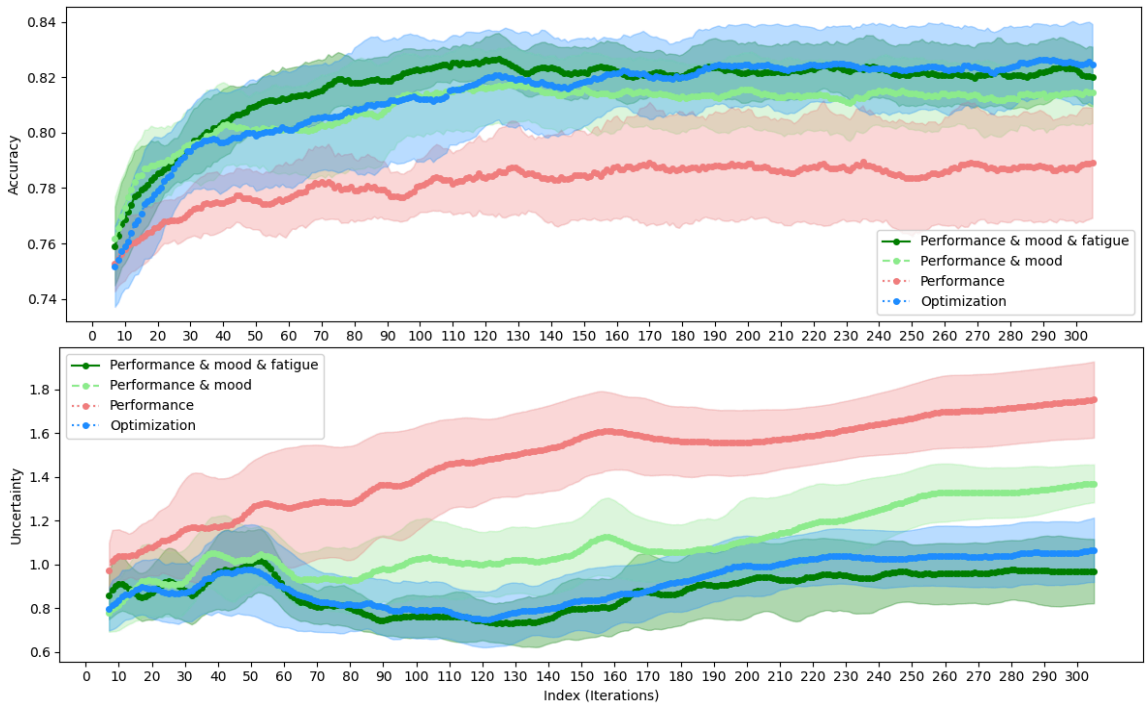

Figure 5.3 - Mean values of model accuracy on full dataset while training, and uncertainty on queried instances for titanic survival prediction dataset with fatigue levels worsening performance by 2%

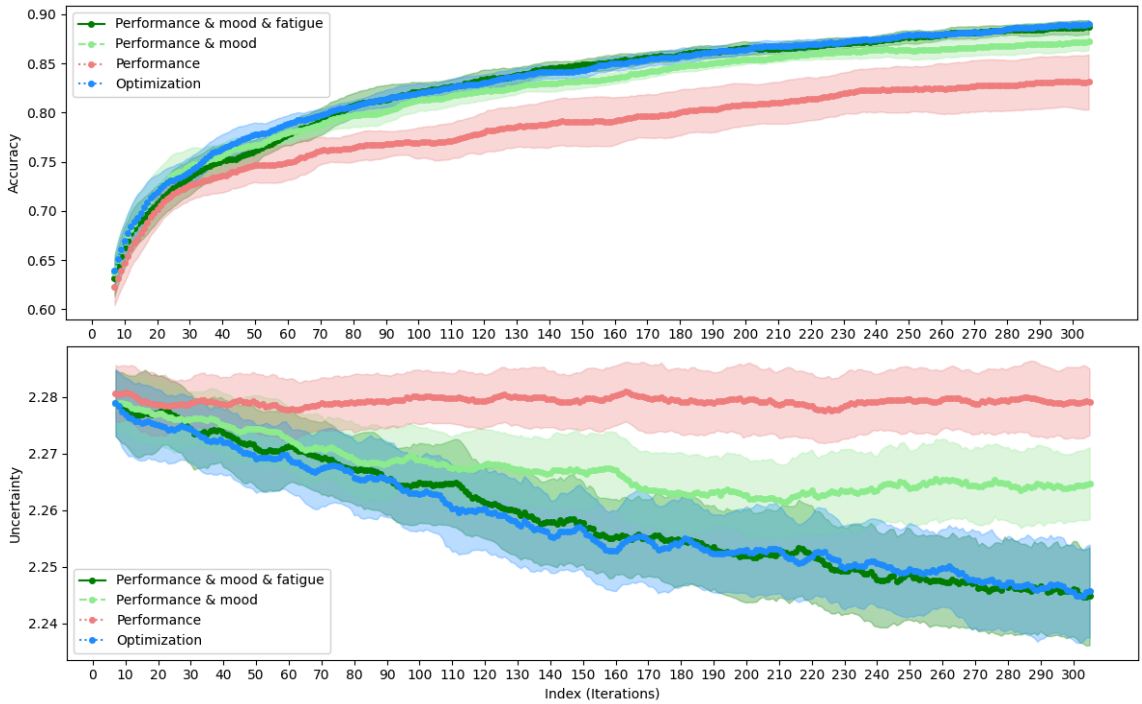

Figure 5.4 - Mean values of model accuracy on full dataset while training, and uncertainty on queried instances for MNIST digit classification dataset with fatigue levels worsening performance by 2%

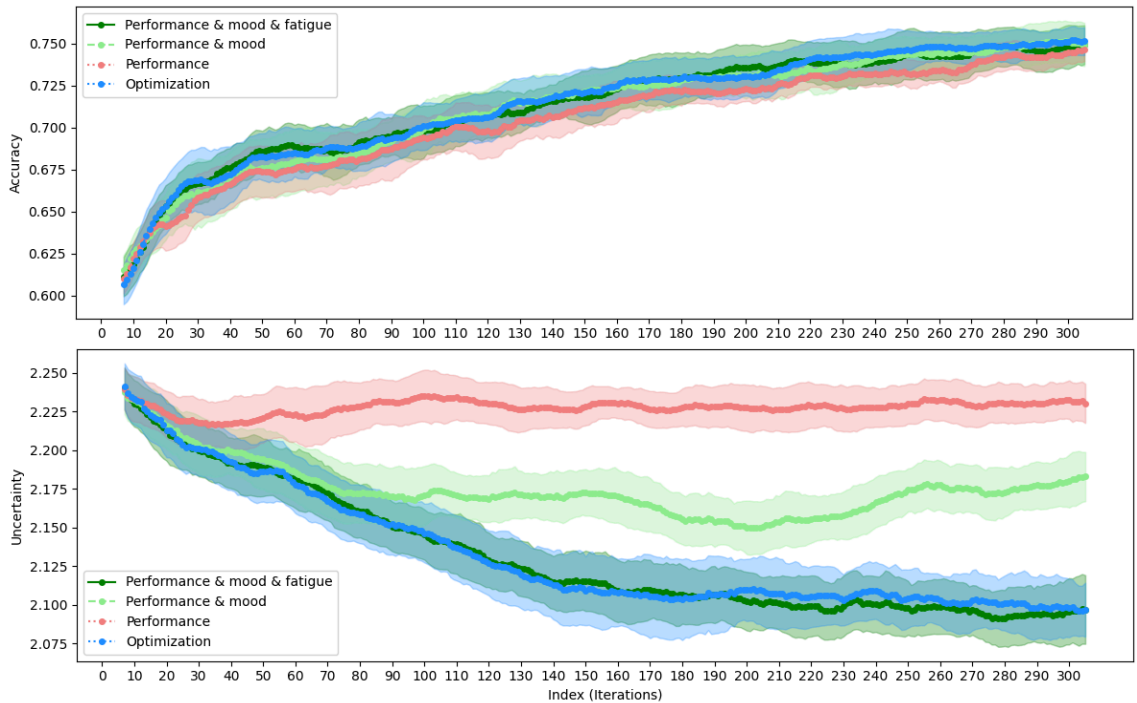

Figure 5.5 - Mean values of model accuracy on full dataset while training, and uncertainty on queried instances for fashion MNIST classification dataset with fatigue levels worsening performance by 2%

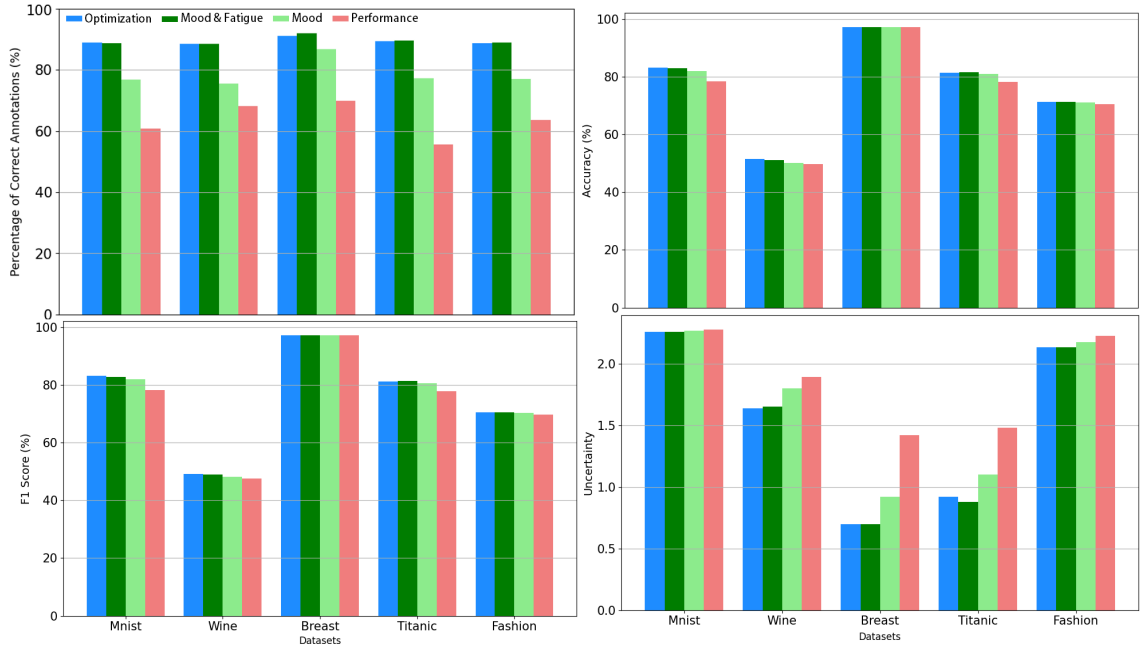

Figure 5.6 - Summary of the results with fatigue levels worsening performance by 2% (percentage of correct annotations, accuracy and F1-score of the model, and uncertainty on queries)

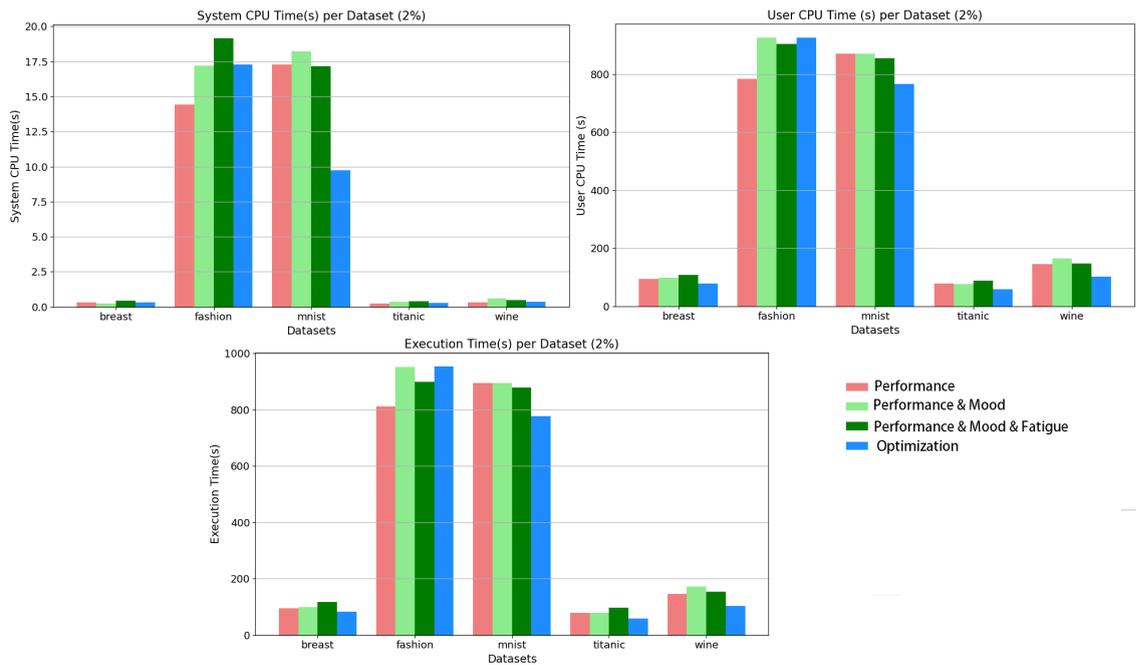

Figure 5.7 - Mean values of system as user CPU times, and execution times per dataset when fatigue levels worsening performance by 2%

We continue by presenting the same graphs in the same order, but for the tests that consider each fatigue level worsening performance by 4%.

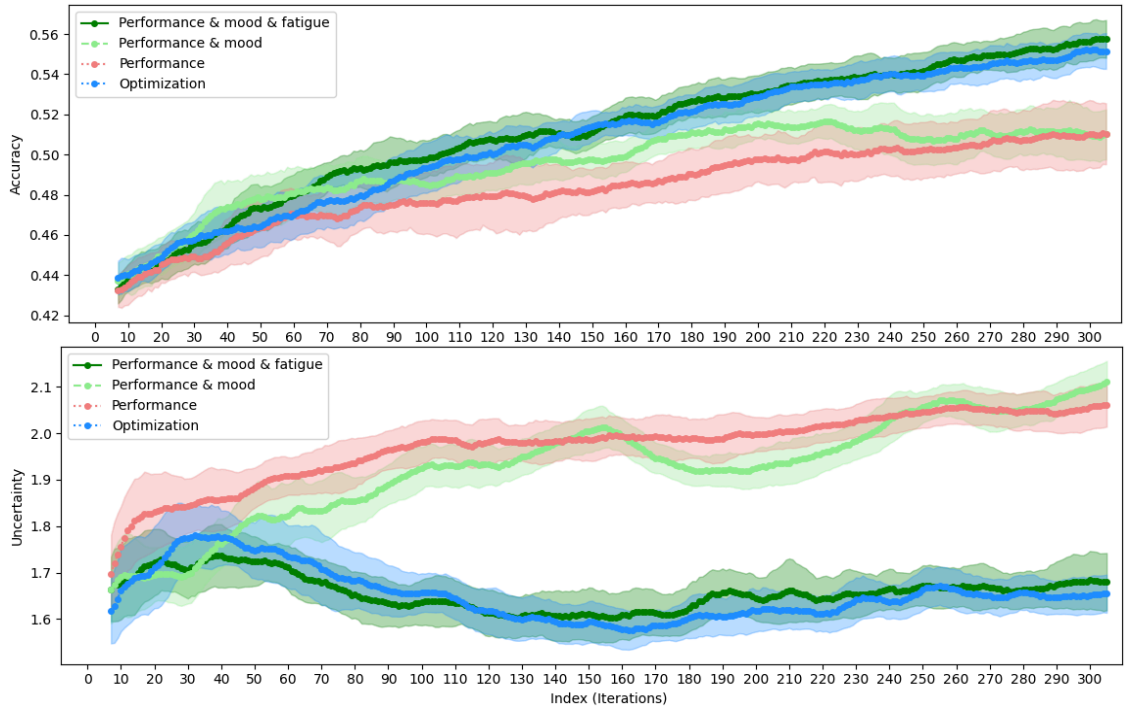

Figure 5.8 - Mean values of model accuracy on full dataset while training, and uncertainty on queried instances for wine quality prediction dataset with fatigue levels worsening performance by 4%

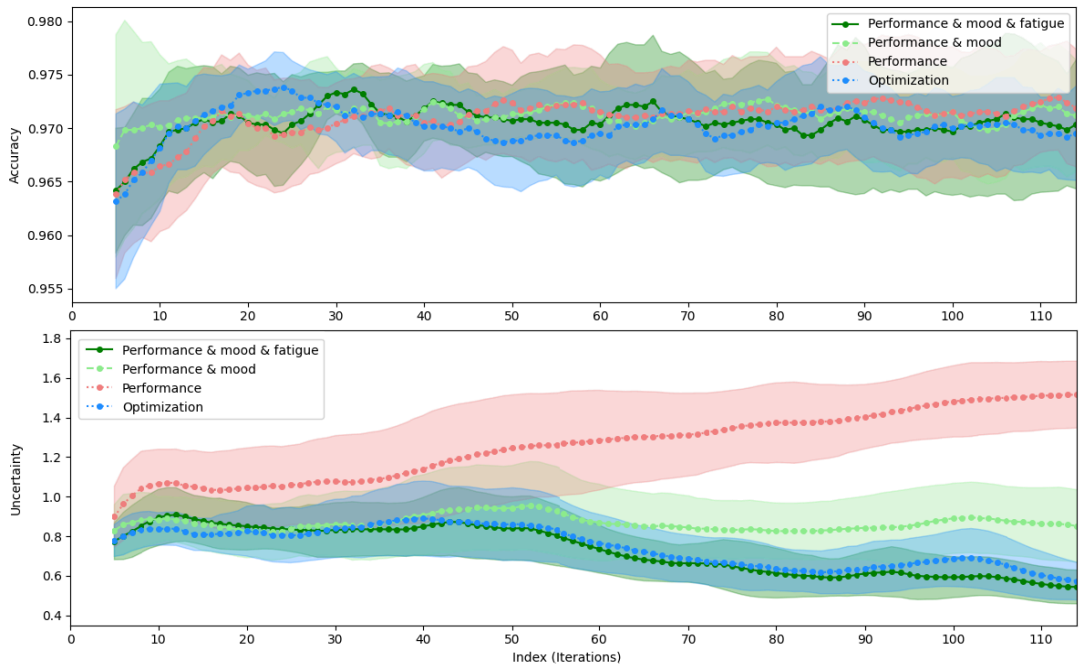

Figure 5.9 - Mean values of model accuracy on full dataset while training, and uncertainty on queried instances for breast cancer detection dataset with fatigue levels worsening performance by 4%

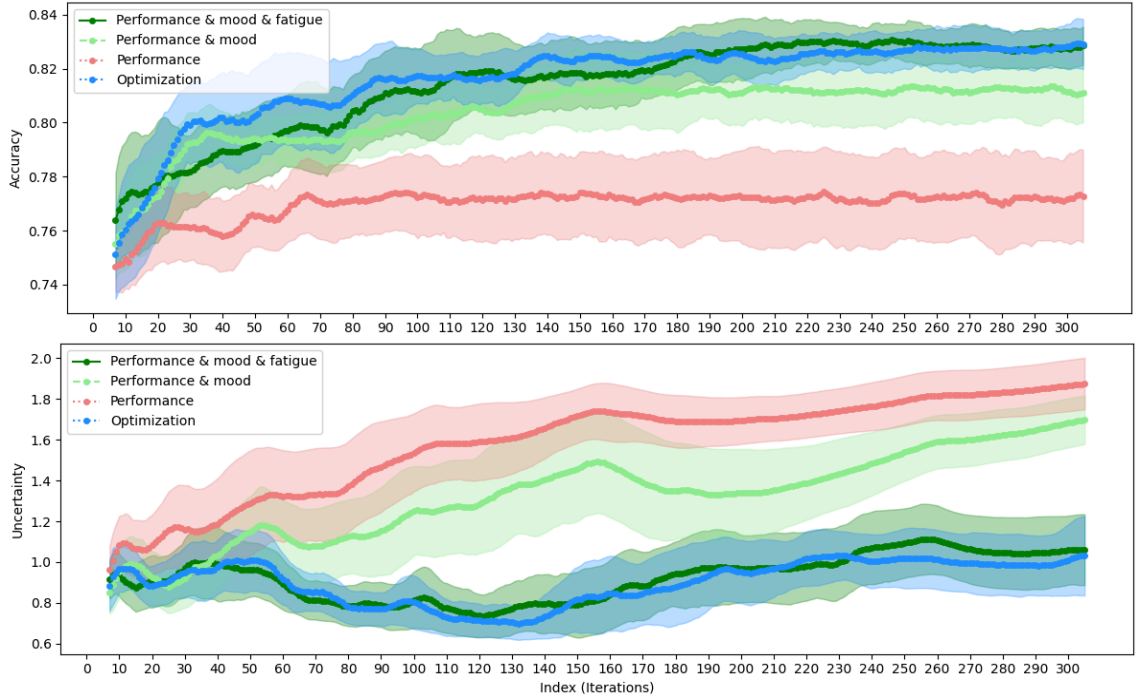

Figure 5.10 - Mean values of model accuracy on full dataset while training, and uncertainty on queried instances for titanic survival prediction dataset with fatigue levels worsening performance by 4%

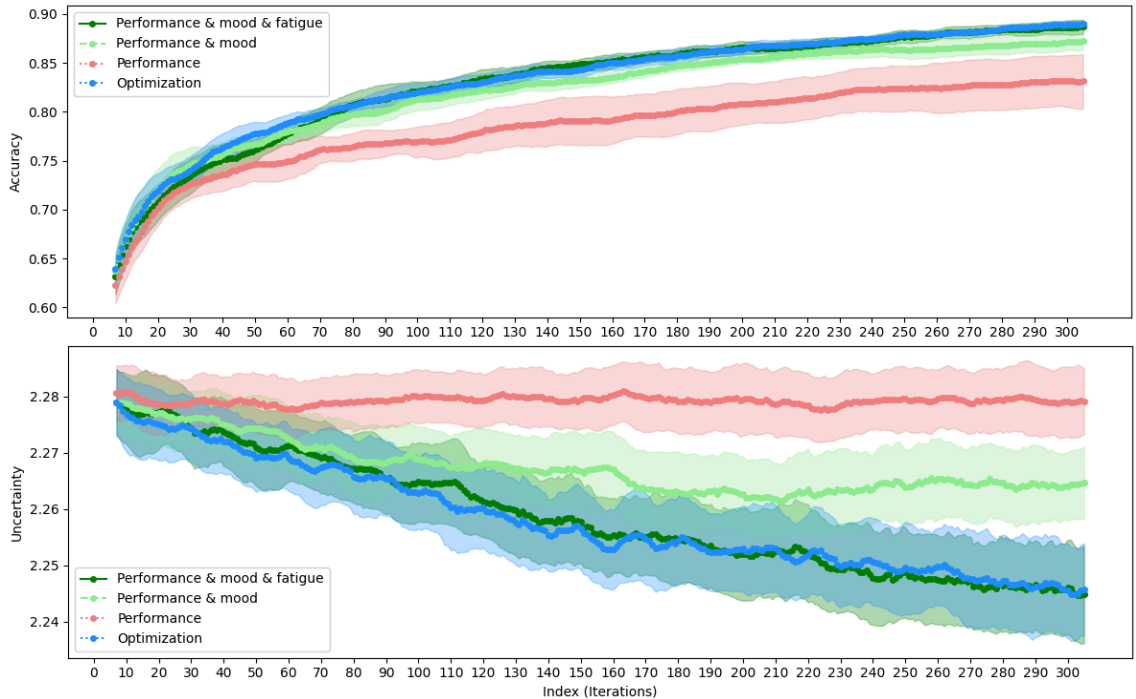

Figure 5.11 - Mean values of model accuracy on full dataset while training, and uncertainty on queried instances for MNIST digit classification dataset with fatigue levels worsening performance by 4%

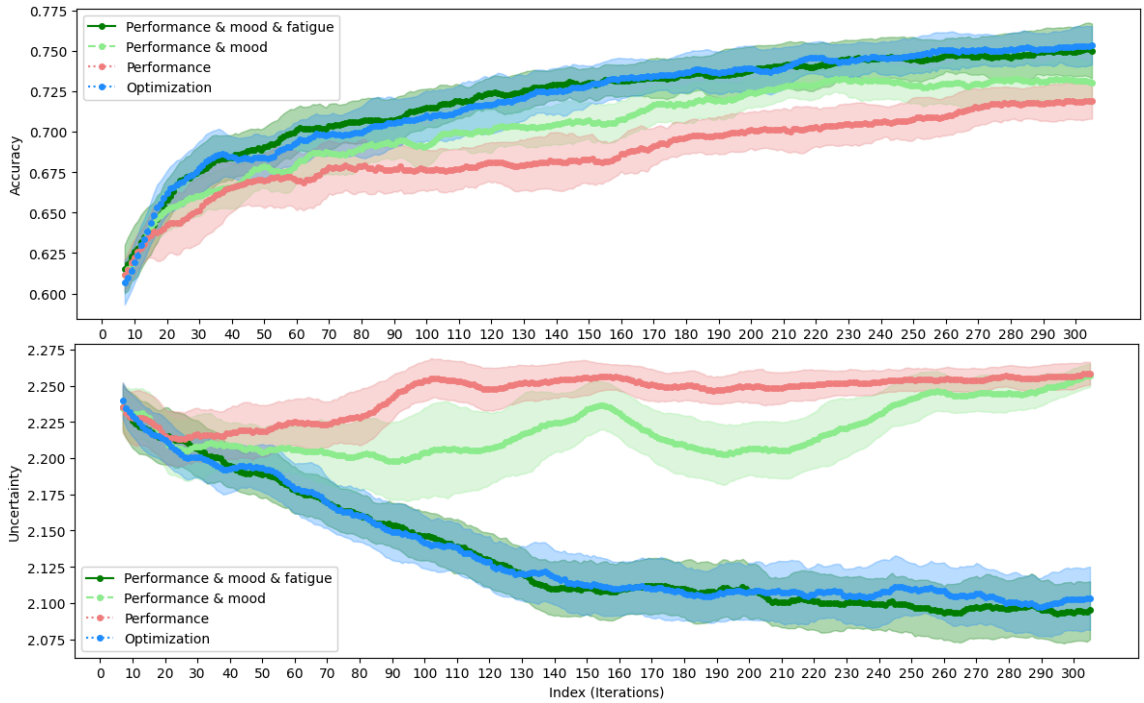

Figure 5.12 - Mean values of model accuracy on full dataset while training, and uncertainty on queried instances for fashion MNIST classification dataset with fatigue levels worsening performance by 4%

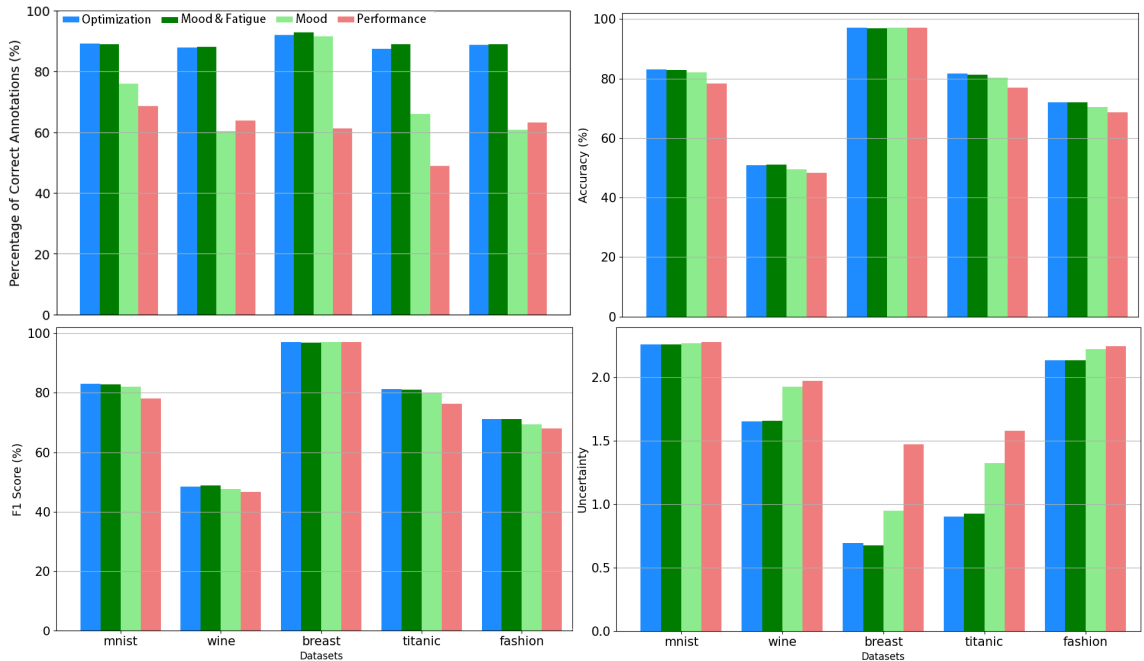

Figure 5.13 - Mean values of model accuracy on full dataset while training, and uncertainty on queried instances for fashion MNIST classification dataset with fatigue levels worsening performance by 4%

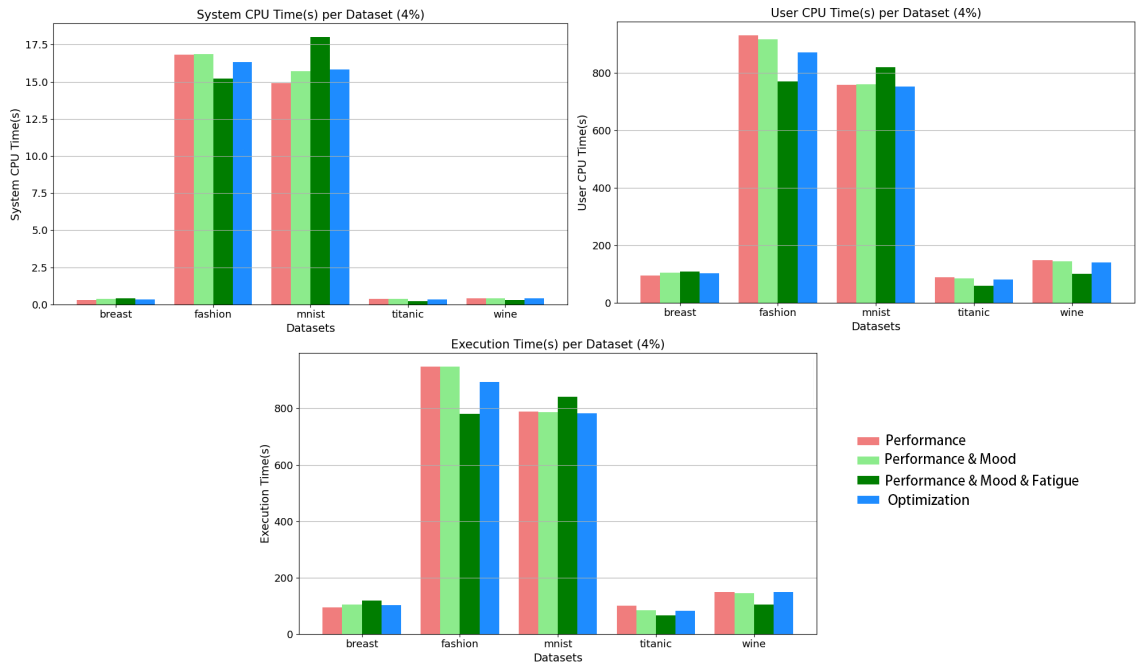

Figure 5.14 - Mean values of model accuracy on full dataset while training, and uncertainty on queried instances for fashion MNIST classification dataset with fatigue levels worsening performance by 4%

## 5.2 Discussion

The figures in the previous section (and the figures found in Appendix B) show that using performance, mood and fatigue levels is better than only considering performance, or performance and mood levels. The only exception is the breast cancer detection dataset, which we will discuss later. This result is expected, as the model is taking into consideration factors that align with the underlying annotation process. For instance, since fatigue levels always affect the annotator in all tests, it is expected that Test 3, that considers this factor, will better predict how the available annotator will perform. This will make those approaches more likely to select the most suitable annotator in that period. Furthermore, the data indicates that the proposed method in Test 3 (our proposed approach) is nearly as effective as the optimized approach, despite datasets and how much fatigue levels are considered in the RS. The more factors are considered in the RS, the more correct annotations are done.

In terms of correct annotations, when considering fatigue levels to worsen accuracy by 2%, our approach (Test 3) is 15.93% better than just considering mood in addition to past accuracy (Test 2), and 36.78% better than the traditional approach (Test 1). When assuming a 4% decrease in performance thanks to fatigue levels, our approach was 24.4% better than considering past accuracy values and mood, and 42.34% better than the traditional approach.

Regarding the mean value of accuracy considering all datasets, when fatigue level decreased performance by 2%, Test 3 had 0.96% better accuracy than Test 2, and 2.49% more accuracy than Test 1. When fatigue level caused a 4% decrease in performance, the accuracy on Test 3 was 1.86% better than Test 2, and 3.65% better than Test 1.

Concerning the mean value of F1-score considering all datasets, with fatigue level discounting 2% in performance, Test 3 had an increase of 0.89% compared to Test 2, and 2.54% increase compared to Test 1. On the other hand, when fatigue level discounted 4% instead, Test 3 was 1.67% better than Test 2, and 3.6% better than Test 1.

When it comes to uncertainty, considering the mean value from all datasets, Test 3 had less 0.242 uncertainty than Test 2, and less 0.442 uncertainty than Test 1 when fatigue level caused a 2% decrease in performance. Furthermore, when this value was 4% instead of 2%, Test 3 had less 0.35 uncertainty compared to Test 2, and less 0.47 uncertainty than Test 1.

In general (excluding breast cancer detection dataset), our proposed approach (Test 3) and the optimized approach (Test 4) achieve a higher F1-score and accuracy faster than Tests 1 and 2. Meanwhile, the uncertainty of the model is also smaller in these two tests. The traditional AL approach (Test 1) achieves in all datasets higher uncertainty of the training process, followed by Test 2, where mood levels are also taken into account. Test 3 and the optimized approach have the smallest uncertainty, with very similar value, through the whole training process.

In the case of the breast cancer detection dataset, its model's uncertainty results are comparable to those of other datasets, exhibiting increased uncertainty in Test 1 (by the biggest margin). However, in the accuracy and F1-score values, every test has similar values. This is true regardless of the consideration of fatigue level in performance. We believe this is caused by how easy the dataset was for the AL model. The accuracy of the model went over 95% since the first iteration in all tests. Since the model is already so good (because of the simplicity of the dataset) with traditional AL technique (Test 1), there is no advantage in using more sophisticated approaches.

Regarding the wine quality dataset, the accuracy and F1-score of the model was close to 50%. The values of Test 3 and optimization show a clear advantage when compared to Test 1 and Test 2. Looking at the values, when each fatigue level affects by 2% an annotator performance, accuracy is 2.23% better in Test 3 compared to Test 2, and 3.33% better in Test 3 compared to Test 1. On the other hand, considering 4% instead of 2% in fatigue level consideration, Test 3 performed 4.77% better than Test 2, and 5.03% better than Test 1. When one fatigue level is considered to decrease performance by 2%, the proposed approach correctly labels 12.9% more instances compared to Test 2, and 20.3% more instances than Test 1. When, instead of 2%, the decrease is 4%, we see an even bigger increase of 26.65% compared to Test 2, and 27.83% compared to Test 1. Since this is the only dataset explored that performed poorly, the data suggests that in those cases it is specially beneficial to use our approach. Its performance (accuracy and F1-score) is very similar to optimization, and there are more correctly annotated instances. The uncertainty, as in all datasets present, has also a bigger advantage when using Test 3, compared to Test 1 or 2. We believe this is evidence of the promising benefits of our proposed approach, in tasks where the model struggles to get good results.

When it comes to the Titanic, MNIST and Fashion MNIST, the accuracy and F1-score values are similar for Test 2, Test 3 and the optimized approach. Despite the similarity, it is still evident that the latter two outperform Test 2, even if by a small margin. We can see a difference in accuracy of 0.4%, 1.2%, and 0.73% respectively between Test 2 and Test 3 considering fatigue level decrease of 2%. Also, a difference in accuracy of 3.13%, 5.17%, and 0.6% respectively between Test 1 and Test 3 in the same context. In the 4% decrease caused by each fatigue level, we see a difference of 1.8%, 1.2%, and 1.33% respectively in accuracy between Test 3 and Test 2. In the same context, we see an accuracy difference of 6.1%, 5.17% and 2.17% between Test 3 and Test 1. This indicates still an advantage of using our proposed approach, when compared to Test 1 or Test 2, but not as significant as in the case of the wine quality assessment dataset.

We believe this is caused by one similarity between these three datasets: how the model considered their difficulty. At the end of the training process, the model's average ranges between around 70-87%. We believe this indicates that, in slightly challenging tasks, even just the use of mood levels in addition to past accuracy values of the annotators can be a good choice for some situations. This suggests that, for example, a team aiming to save money might prefer relying solely on mood levels, accepting a small reduction in accuracy, rather than incurring additional costs to measure fatigue levels for near-optimal accuracy. However, there is still an advantage for choosing our proposed approach, instead of just considering mood values combined with past accuracy of the annotators.

Moreover, comparing the usage of 2% or 4% decrease performance due to an increase in fatigue level, using 4% instead of 2% shows to have bigger advantages when comparing our proposed approach with Test 2 and Test 1. Using 4%, on the other hand, we can see uncertainty of the model on Test 2 be higher, coming close to the values of Test 1. While it is not the purpose of the study to explore the difference, but rather to present them while no empirical data on a realistic value is found, we can conclude that the more fatigue affects performance, the more our approach is beneficial.

CPU and execution times show inconclusive findings. There is no conclusion worth mentioning, despite the obvious correlation with the dimensionality of the datasets, and the data type of MNIST and fashion MNIST datasets. There were also differences between the results of the three batches of annotators that were not worth noting.

The results of this research reveal that our proposed approach is in general a better option than just considering past accuracy of the annotators. This is true for the four metrics analyzed (model accuracy, F1-score and uncertainty through training, as well as correct annotations performed by the annotators). The proposed method is very similar to the usage of an optimization-based approach. The advantages in accuracy and F1-score are fairly small, but the difference in correct annotation performed and uncertainty of the model in the queried labels through training is encouraging. It indicates a promising future of solving the open question in AL of, not only allowing and expecting annotators to be wrong sometimes, but to try to predict by how much as best as cognitive research allows.

# Chapter 6
# Conclusion

The presented research shows that the suggested approach for active learning (AL), which takes cognitive aspects into account to enhance the prediction of annotators' performance during labeling, has a promising future in the field. It contributes to solving the open challenge of how AL often considers only annotators' past accuracy to select query-annotator pairs, neglecting human cognitive factors that can influence their performance. The proposed method includes the use of a recommendation system (RS) to select the best query-annotator pair for the labeling process in an AL task. The RS takes into account all the available annotators' mood and fatigue levels at a given time period, as well as their past accuracy (overall and label based) to more correctly predict their performance while labeling a given instance. Using this approach in different datasets with different characteristics (sizes, fields, modality, number of labels and objective difficulties), while considering the effect of cognitive and emotional states that influence human's performance, allowed the exploration and comparison of the results.

The existing literature was the basis for our decision on how mood impacts performance, however, no studies were identified that provided quantitative measures of fatigue's effects on performance. For that reason, we tested two different values. We repeated the tests considering a unit increase of fatigue level lead to 2% or 4% worse performance in two consecutive periods of work in the same day. We explored the results in both, as there is yet no data to safely assume one of the values over the other. We saw bigger advantages in using our approach when we considered a 4% decrease in accuracy, compared to only using 2%.

To test the proposed approach we used five datasets: wine quality prediction, breast cancer detection, titanic survival prediction, MNIST digit classification, fashion MNIST classification. These datasets vary in characteristics, such as modality, size and fields. The use of multiple different dataset allows for a more detailed examination of the results and usability of the proposed approach. The study also compared these results with an optimized approach, used to have a baseline for more objective comparison in each dataset. This optimization was based on the function that simulates how the annotators actually label the data.

The Knowledge-based RS developed in this study for selecting the best query-annotator pairs for AL proved to be better than traditional AL methods, approaching the effectiveness of an optimized approach. The annotators correctly labeled 36.78% more instances than the traditional approach, when fatigue levels were considered to worsen performance by 2%, and 42.34% when we tested for a 4% decrease instead. Accuracy and F1-score also presented improvements in the proposed approach. When considering a 4% decrease in performance due to each fatigue level, accuracy and F1-score had a 3.65% and 3.6% increase compared to the traditional AL approach, respectively. In the case of a 2% decrease, there was a 2.49% and 2.54% increase compared to the traditional AL approach (that considers only past annotator accuracy), respectively.

The results suggest that this approach is especially useful in cases where traditional AL struggles to achieve better than poor results (around 40-50% accuracy). Its benefits are also present for datasets resulting in around 70-80% accuracy with traditional AL. On the other hand, in cases where traditional AL already achieved very high accuracy (above 95%), the proposed approach only seems to decrease the uncertainty of the model while training, but achieves similar results in performance. Additionally, in all datasets, regardless of difficulty for a traditional AL model, the proposed methodology presented smaller uncertainty of the queried instances during training, as well as more correctly labeled instances by the annotators.

Future work should explore different weights in the RS, exploring how different values change the effectiveness of the proposed approach, since this study bases itself on empirical data, and creates arbitrary conservatory values when they are missing. It is crucial to assess different choices of values, and observe how they affect the proposed approach. It is also important to test in more different batches of annotators, to investigate the utility of study across annotation teams with differing levels of expertise and quality requirements.

Furthermore, it is also important to consider how the study translates in a real world setting, with real annotators. It is important to assess mood influence on performance in specifically annotation tasks (so that it is not needed to use a conservative value based on similar, but different tasks). Furthermore, it is crucial to also consider different methods for accessing fatigue and mood levels. Beyond the self reported values, more objective methods can be used. While studying fatigue and mood influence on humans performance in annotation tasks already results in empirical data valuable to use in this approach, using physiological/biometrics monitoring allows for more precise measurements, such as size of pupils and heart rate variability [Dzedzickis et al., 2020, Karim et al, 2024, Ramírez-Moreno et al., 2021].

Additionally, the integration of cognitive factors of all available annotators in the RS might have scalability problems that should be further investigated with more annotators per batch considered.

Although the study is only done in simulated annotators, and some values are arbitrary due to inexistent literature on the field, our approach has proven to be a promising approach in the AL field. While the increases on accuracy and F1-score might be low, the study shows a substantial improvement in uncertainty of the model, and the amount of correct annotations performed. This approach contributed to the development of techniques that include cognitive factors to better predict annotators' performance, choosing the best query-annotator pairs each time. It begins to explore the potential of integrating cognitive-based aspects with the AL field, achieving a more realistic HitL framework. This approach not only considers the worker's well-being but also leverages it to get the least amount of human error, and the best model results.

Chapter 6 - Conclusion

# Appendix A
# Annotator batches

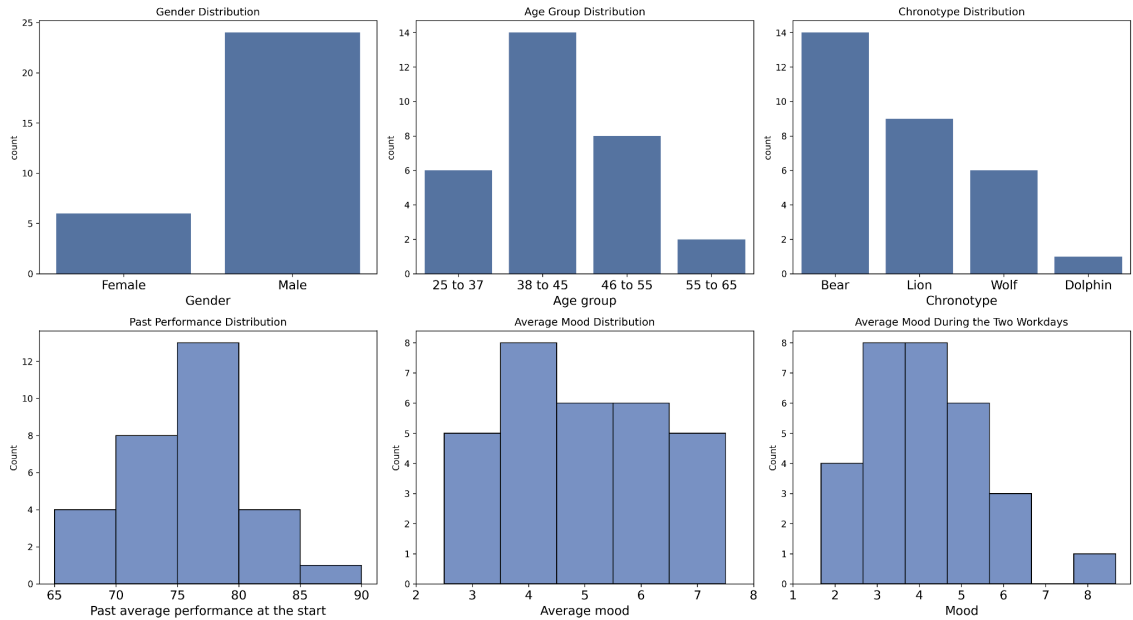

Figure A.1 - Distribution of features in annotator batch 1

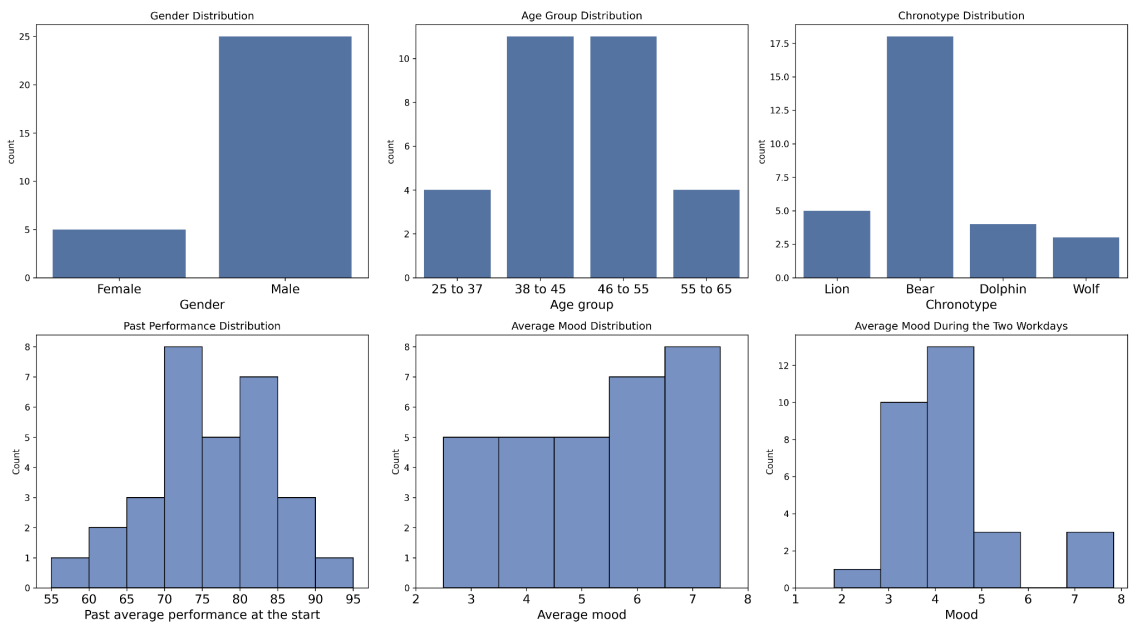

Figure A.2 - Distribution of features in annotator batch 2

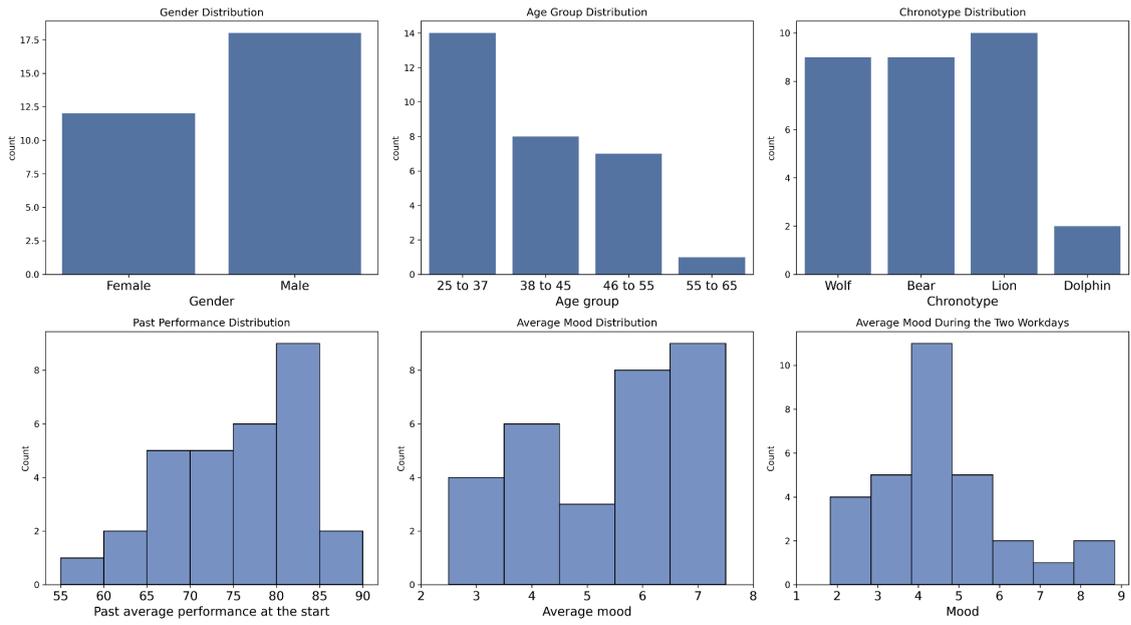

Figure A.3 - Distribution of features in annotator batch 3

# Appendix B
# Additional results

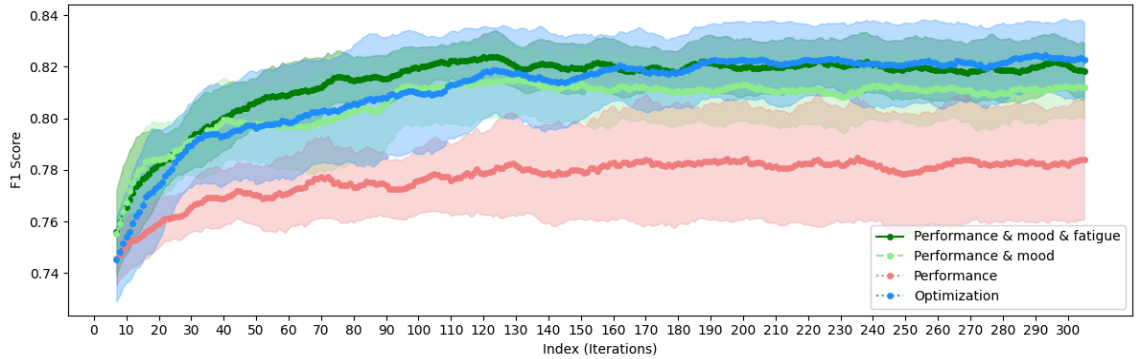

Figure B.1 -Mean values of model F1-score on full dataset while training for wine quality prediction with fatigue levels worsening performance by 2%

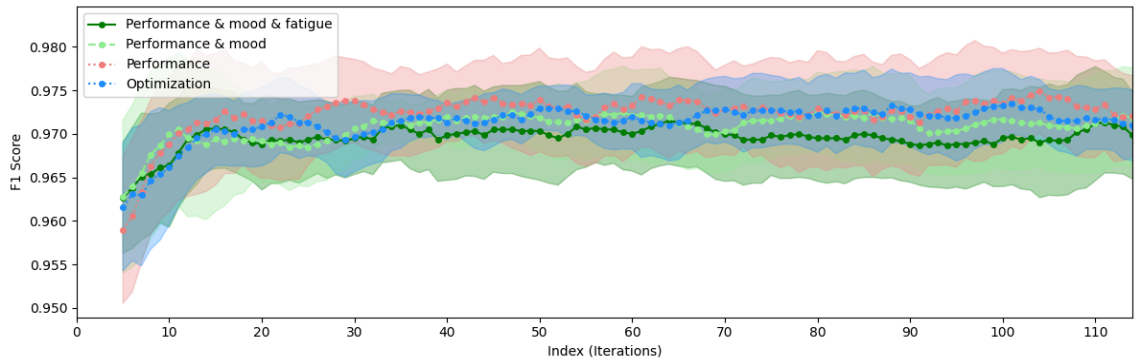

Figure B.2 - Mean values of model F1-score on full dataset while training for breast cancer detection with fatigue levels worsening performance by 2%

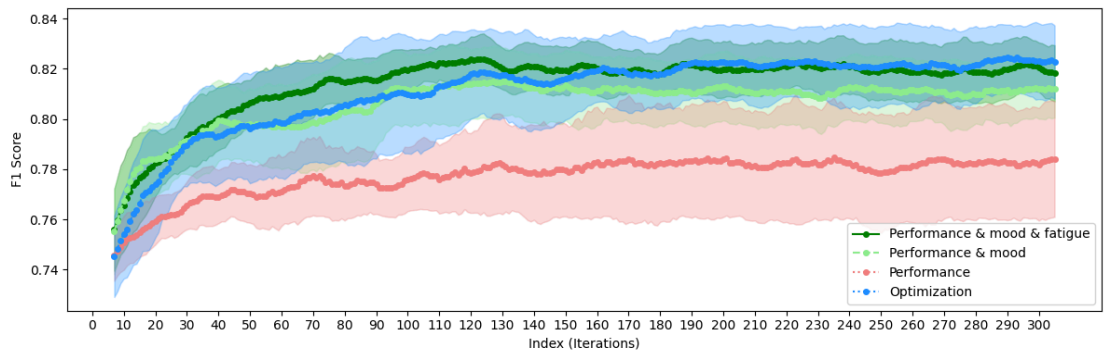

Figure B.3 - Mean values of model F1-score on full dataset while training for titanic survival prediction with fatigue levels worsening performance by 2%

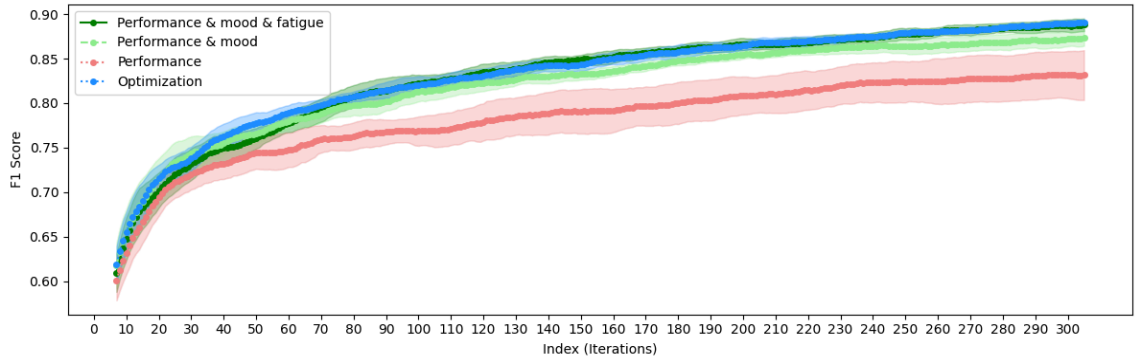

Figure B.4 - Mean values of model F1-score on full dataset while training for MNIST digit classification with fatigue levels worsening performance by 2%

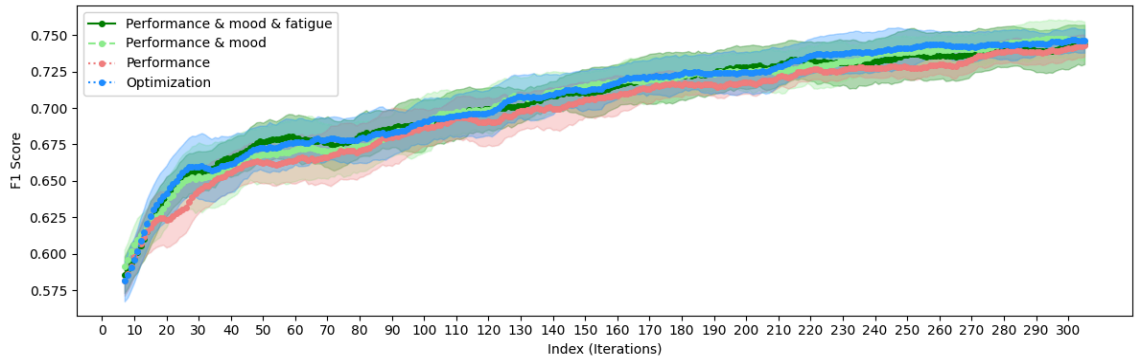

Figure B.5 - Mean values of model F1-score on full dataset while training for fashion MNIST classification with fatigue levels worsening performance by 2%

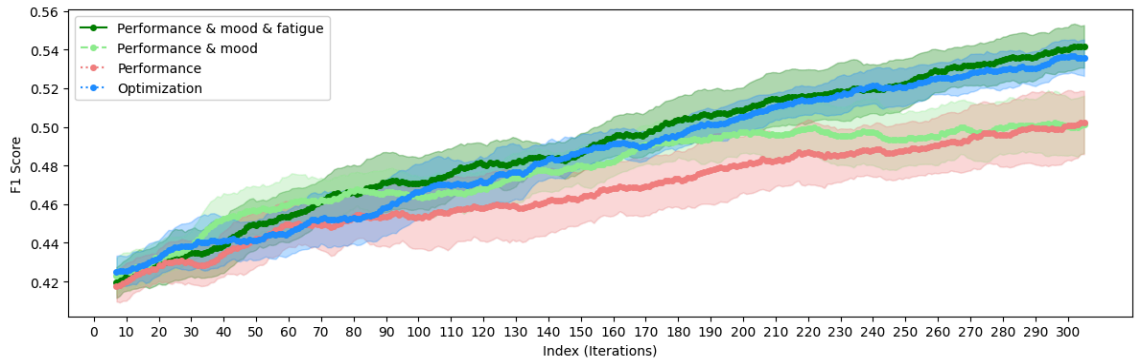

Figure B.6 - Mean values of model F1-score on full dataset while training for wine quality prediction dataset with fatigue levels worsening performance by 4%

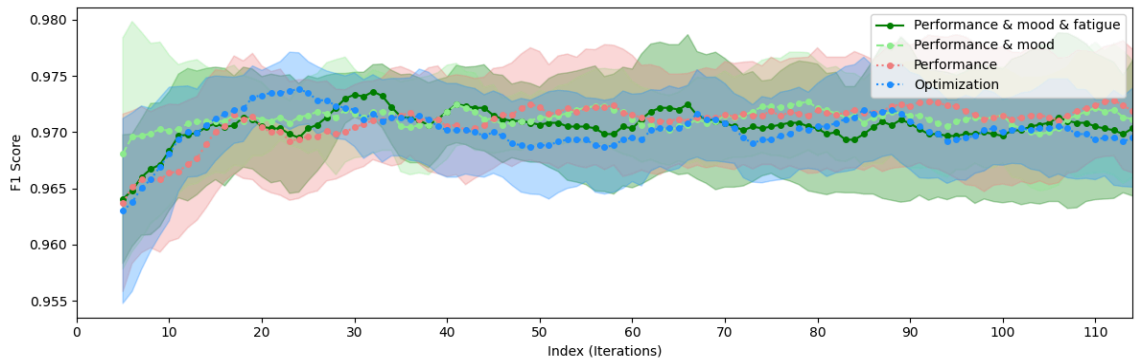

Figure B.7 - Mean values of model F1-score on full dataset while training for breast cancer detection with fatigue levels worsening performance by 4%

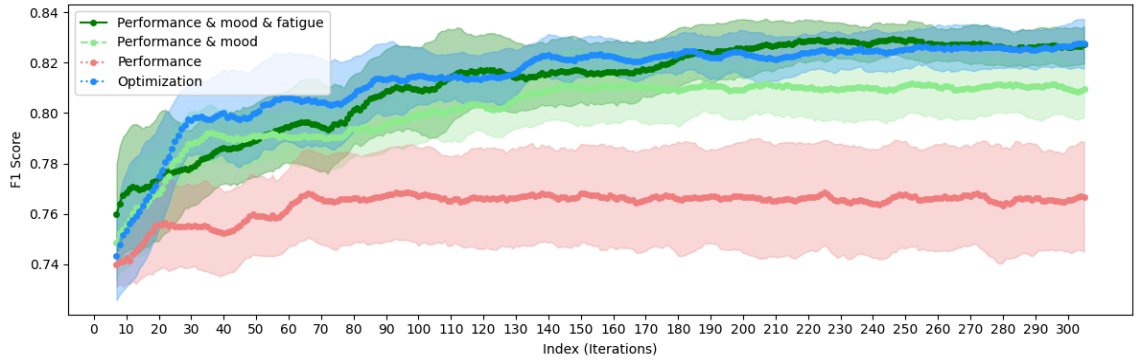

Figure B.8 - Mean values of model F1-score on full dataset while training for titanic survival prediction dataset with fatigue levels worsening performance by 4%

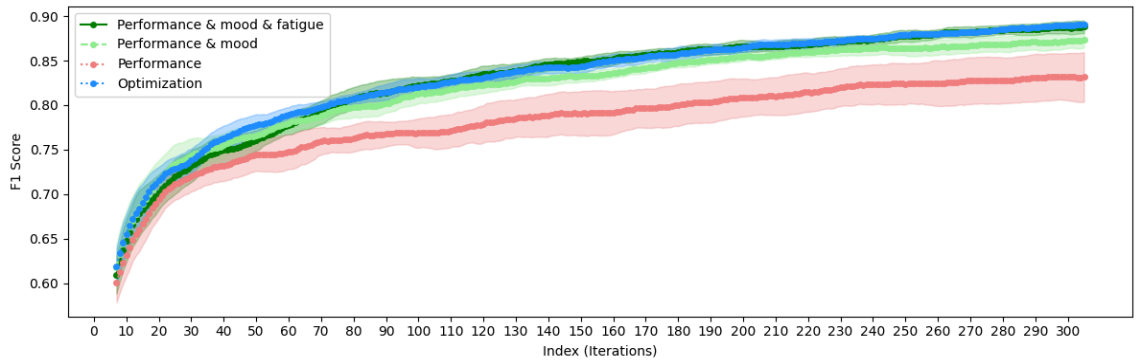

Figure B.9 - Mean values of model F1-score on full dataset while training for MNIST digit classification with fatigue levels worsening performance by 4%

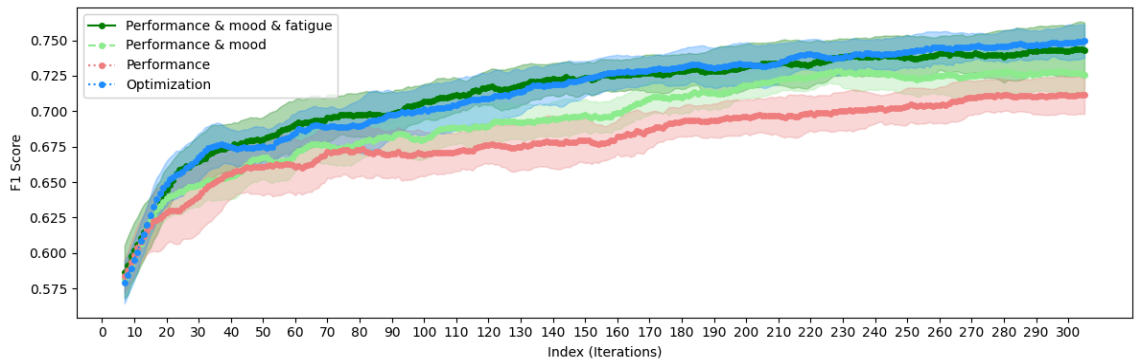

Figure B.10 - Mean values of model F1-score on full dataset while training for fashion MNIST classification with fatigue levels worsening performance by 4%